%% file: main.tex
\documentclass{article}

\PassOptionsToPackage{round}{natbib}

\usepackage[english]{babel}
\usepackage[utf8]{inputenc}
\usepackage[T1]{fontenc}
\usepackage{amsmath}
\usepackage{amsfonts} 
\usepackage{amssymb}

\usepackage[colorinlistoftodos, textsize=tiny]{todonotes}
\usepackage{soul}
\usepackage{framed}

\usepackage[normalem]{ulem}
\usepackage{phaistos}
\usepackage{wrapfig}
\usepackage{microtype}
\usepackage{graphicx}
\usepackage{subcaption}
\usepackage{booktabs} 
\usepackage{hyperref}

\usepackage[preprint]{nips_2018}

\title{A Walk with SGD}

\author{Chen Xing\thanks{Equal Contribution}\, \thanks{College of Computer and Control Engineering, Nankai University, xingchen1113@gmail.com}\, \thanks{MILA, Université de Montréal}\,\,\,\,Devansh Arpit\footnotemark[1]\, \thanks{MILA, Université de Montréal, devansharpit@gmail.com}\,\,\,\, Christos Tsirigotis\thanks{Aristotle University of Thessaloniki, Greece}\,  \footnotemark[3]\,\,\,\, Yoshua Bengio\footnotemark[3]\, \thanks{CIFAR Senior Fellow}}

\begin{document}
\maketitle

\begin{abstract}
We present novel empirical observations regarding how stochastic gradient descent (SGD) navigates the loss landscape of over-parametrized deep neural networks (DNNs). These observations expose the qualitatively different roles of learning rate and batch-size in DNN optimization and generalization. Specifically we study the DNN loss surface along the trajectory of SGD by interpolating the loss surface between parameters from consecutive \textit{iterations} and tracking various metrics during training. We find that the loss interpolation between parameters before and after each training iteration's update is roughly convex with a minimum (\textit{valley floor}) in between for most of the training. Based on this and other metrics, we deduce that for most of the training update steps, SGD moves in valley like regions of the loss surface by jumping from one valley wall to another at a height above the valley floor. This 'bouncing between walls at a height' mechanism helps SGD traverse larger distance for small batch sizes and large learning rates which we find play qualitatively different roles in the dynamics. While a large learning rate maintains a large height from the valley floor, a small batch size injects noise facilitating exploration. We find this mechanism is crucial for generalization because the valley floor has barriers and this exploration above the valley floor allows SGD to quickly travel far away from the initialization point (without being affected by barriers) and find flatter regions, corresponding to better generalization. 

\end{abstract}

\section{Introduction}
Deep neural networks (DNNs) trained with algorithms based on stochastic gradient descent (SGD) are able to tune the parameters of massively over-parametrized models to reach small training loss with good generalization despite the existence of numerous bad minima. This is especially surprising given DNNs are capable of overfitting random data with almost zero training loss \cite{zhang2016understanding}. This behavior has been studied by \citet{arpit2017closer, advani2017high} where they suggest that deep networks generalize well because they tend to fit simple functions over training data before overfitting noise. It has been further discussed that model parameters that are in a region of flatter minima generalize better \cite{hochreiter1997flat, keskar2016large, wu2017towards}, and that SGD finds such minima when used with small batch size and large learning rate \cite{ keskar2016large, jastrzkebski2017three, smith2017don, chaudhari2017stochastic}. These recent papers frame SGD as a stochastic differential equation (SDE) under the assumption of using small learning rates. A main result of these papers is that the SDE dynamics remains the same as long as the ratio of learning rate to batch size remains unchanged. \textit{However, this view is limited due to its assumption and ignores the importance of the structure of SGD noise (i.e., the gradient covariance) and the qualitative roles of learning rate and batch size, which remain relatively obscure.}

On the other hand, various variants of SGD have been proposed for optimizing deep networks with the goal of addressing some of the common problems found in the high dimensional non-convex loss landscapes (Eg. saddle points, faster loss descent etc). Some of the popular algorithms used for training deep networks apart from vanilla SGD are SGD with momentum \cite{polyak1964some, sutskever2013importance}, AdaDelta \cite{zeiler2012adadelta}, RMSProp \cite{tieleman2012lecture}, Adam \cite{kingma2014adam} etc. However, for any of these methods, currently there is little theory proving they help in improving generalization in DNNs (which by itself is currently not very well understood), although there have been some notable efforts (Eg. \citet{hardt2015train,kawaguchi2017generalization}). \textit{This raises the question of whether optimization algorithms that are designed with the goal of solving the aforementioned high dimensional problems also help in finding minima that generalize well, or put differently, what attributes allow optimization algorithms to find such good minima in the non-convex setting.}

We take a step towards answering the above two questions in this paper for SGD (without momentum) through qualitative experiments. The main tool we use for studying the DNN loss surface along SGD's path is to interpolate the loss surface between parameters before and after each training update and track various metrics. Our findings about SGD's trajectory can be summarized as follows:

1. We observe that the loss interpolation between parameters before and after each \textit{iteration's} update is roughly convex with a minimum (\textit{valley floor}) in between. Thus, we deduce that SGD bounces off walls of a \textit{valley-like-structure} at a height above the floor. 

2. Learning rate controls the height at which SGD bounces above the valley floor while batch size controls gradient stochasticity which facilitates exploration (visible from larger parameter distance from initialization for small batch-size). In this way, learning rate and batch size exhibit different qualitative roles in SGD dynamics.\footnote{This implies that except when using a reasonably small learning rate (which would make the SDE approximation hold), the effect of small batch size with a certain learning rate cannot be achieved by using a large batch size with a proportionally large learning rate (observed by \citet{goyal2017accurate}).}

3. The valley floor along SGD's path has many ups and downs (barriers) which may hinder exploration. Thus using a large learning rate helps avoid encountering barriers along SGD's path by maintaining a large height above valley floor (thus moving over the barriers instead of crossing\footnote{A \textbf{barrier is crossed} when we see a point in the parameter space interpolated between the points just before and after an update step, such that the loss at the barrier point is higher than the loss at both the other points.} them). 


Experiments are conducted on multiple data sets, architectures and hyper-parameter settings. The findings mentioned above hold true on all of them. We further find that stochasticity in SGD induced by mini-batches is needed both for better optimization and generalization. Conversely, artificially added isotropic noise in the absence of mini-batch induced stochasticity is bad for DNN optimization. We also discuss some striking similarities between our empirical findings about SGD's trajectory in DNNs and classical optimization theory in the quadratic setting.

\vspace{-5pt}
\section{Background and Related Work}

Various algorithms have been proposed for optimizing deep neural networks that are designed from the view point of tackling various high dimensional optimization problems like oscillation during training (SGD with momentum \cite{polyak1964some}), oscillations around minima (Nesterov momentum \cite{nesterov1983method, sutskever2013importance}), saddle points \cite{dauphin2014identifying}, automatic decay of the learning rate (AdaDelta \cite{zeiler2012adadelta}, RMSProp \cite{tieleman2012lecture} and ADAM \cite{kingma2014adam}), etc. 
However, currently there is insufficient theory to understand what kind of minima generalize better although empirically it has been observed that wider minima (that can be quantified by low Hessian norm) seem to have better generalization \cite{ keskar2016large, wu2017towards, jastrzkebski2017three} due to their low complexity and are more likely to be reached under random initialization given their larger volumes \cite{wu2017towards}. 
\textit{This argument raises the question of whether the intuitions behind the designs of the various optimization algorithms are really the reasons behind their success in deep learning or there are other underlying mechanisms that make them successful.}

To understand this aspect better, a number of (mostly) recent papers study SGD as a stochastic differential process \cite{kushner2003stochastic, mandt2017stochastic, chaudhari2017stochastic, smith2017understanding, jastrzkebski2017three, li2015stochastic} under the assumption (among others) that the learning rate is reasonably small. Broadly, these papers show that the stochastic fluctuation in the stochastic differential equation simulated by SGD is governed by the ratio of learning rate to batch size. Hence according to this theoretical framework, the training dynamics of SGD should remain roughly identical when changing learning rate and batch size by the same factor. However, given DNNs (especially Resnet \cite{he2016deep} like architectures) are often trained with quite large learning rates, the small learning rate assumption may be a pitfall of this theoretical framework\footnote{For instance \citet{goyal2017accurate} investigate that increasing learning rate linearly with batch size helps to a certain extent but breaks down for very large learning rates.}. But this theory is nonetheless useful since learning rates do attain small values during training due to annealing or adaptive scheduling, so this framework may indeed apply during parts of training. \textit{In this paper we attempt to go beyond these analyses to study the different qualitative roles of the noise induced by large learning rate versus the noise induced by a small batch size.}
\begin{wrapfigure}{r}{0.48\textwidth}
\vspace{-10pt}
\begin{subfigure}{0.45\columnwidth}
  \centering
  \includegraphics[scale=0.6,trim=0.15in 0.35in 0.in 0.1in,clip]{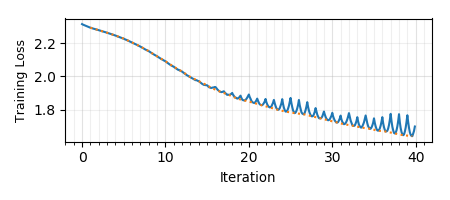}
\end{subfigure}\hspace{0.2\textwidth}
\begin{subfigure}[t]{0.45\columnwidth}
  \centering
    \includegraphics[scale=0.6,trim=0.15in 0.35in 0.in 0.in,clip]{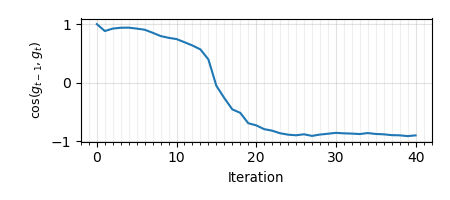}
\end{subfigure}\hspace{0.1\textwidth}
\begin{subfigure}{0.45\columnwidth}
  \centering
  \includegraphics[scale=0.6,trim=0.15in 0.1in 0.in 0.in,clip]{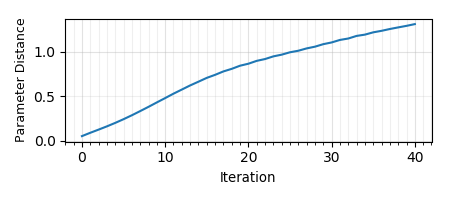}
\end{subfigure}
\caption{Plots for VGG-11 Epoch 1 trained using full batch \textbf{Gradient Descent} (GD) on CIFAR-10. \textbf{Top}: Training loss for the $1st$ $40$ iterations of training. Between the training loss at every consecutive iteration (vertical gray lines), we uniformly sample $10$ points between the parameters before and after a training update and calculate the loss at these points. Thus we take a slice of the loss surface between two iterations. These loss values are plotted between every consecutive training loss value from training updates. The dashed orange line connects the minimum of the loss interpolation between consecutive iterations (this minimum denotes the valley floor along the interpolation). \textbf{Middle}: Cosine of the angle between gradients from two consecutive iterations. \textbf{Bottom}: Parameter distance from initialization. \textbf{Gist}: The loss interpolation between consecutive iterations have a minimum for iterations where cosine is highly negative (close to $-1$ after around $20$ iterations meaning the consecutive gradients are almost along opposite directions), suggesting the optimization is oscillating along the walls of a valley like structure. The valley floor reduces monotonously. \label{fig:loss_interpolation_gd_vgg11_cifar10_boom}}
\vspace{-30pt}
\end{wrapfigure}

There have also been work that consider SGD as a diffusion process where SGD is running a Brownian motion in the parameter space. \citet{li2017batch} hypothesize this behavior of SGD and theoretically show that this diffusion process would allow SGD to cross barriers and thus escape sharp local minima. The authors use this theoretical result to support the findings of \citet{keskar2016large} who find that SGD with small mini-batch find wider minima. \citet{hoffer2017train} on the other hand make a similar hypothesis based on the evidence that the distance moved by SGD from initialization resembles a diffusion process, and make a similar claim about SGD crossing barriers during training. \textit{Contrary to these claims, we find that interpolating the loss surface traversed by SGD on a per iteration basis suggests SGD almost never crosses any significant barriers for most of the training.}

There is also a long list of work towards understanding the loss surface geometry of DNNs from a theoretical standpoint. \citet{dotsenko1995introduction, amit1985spin, choromanska2015loss} show that under certain assumptions, the DNN loss landscape can be modeled by a spherical spin glass model which is well studied in terms of its critical points. \citet{safran2016quality} show that under certain mild assumptions, 
the initialization is likely to be such that there exists a continuous monotonically decreasing path from the initial point to the global minimum. \citet{freeman2016topology} theoretically show that for DNNs with rectified linear units (ReLU), the level sets of the loss surface become more connected as network over-parametrization increases. This has also been justified by \citet{sagun2017empirical} who show that the hessian of deep ReLU networks is degenerate when the network is over-parametrized and hence the loss surface is flat along such degenerate directions. \citet{goodfellow2014qualitatively} empirically show that the convex interpolation of the loss surface from the initialization to the final parameters found by optimization algorithms do not cross any significant barriers, and that the landscape of loss surface near SGD’s trajectory has a valley-like 2D projection. \textit{Broadly these studies analyze DNN loss surfaces (either theoretically or empirically) in isolation from the optimization dynamics.}

In our work we do not study the loss surface in isolation, but rather analyze it through the lens of SGD. In other words, we study the DNN loss surface along the trajectory of SGD and track various metrics while doing so, from which we deduce both how the landscape relevant to SGD looks like, and how the hyperparameters of SGD (learning rate and batch size) help SGD maneuver through it.

\section{A Walk with SGD}
\label{the_walk}
\begin{wrapfigure}{r}{0.48\textwidth}
\vspace{-35pt}
 \centering
\begin{subfigure}{0.45\columnwidth}
  \centering
  \includegraphics[scale=0.6,trim=0.15in 0.35in 0.in 0.1in,clip]{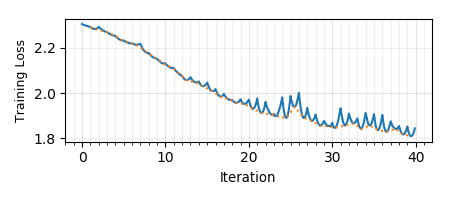}
\end{subfigure}\hspace{0.2\textwidth}
\begin{subfigure}{0.45\columnwidth}
  \centering
   \includegraphics[scale=0.6,trim=0.15in 0.35in 0.in 0.in,clip]{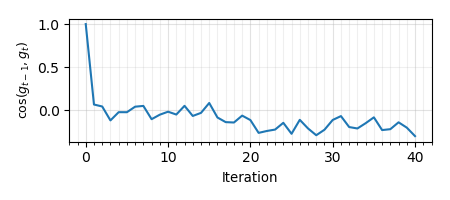}
\end{subfigure}\hspace{0.1\textwidth}
\begin{subfigure}{0.45\columnwidth}
  \centering
  \includegraphics[scale=0.6,trim=0.15in 0.1in 0.in 0.in,clip]{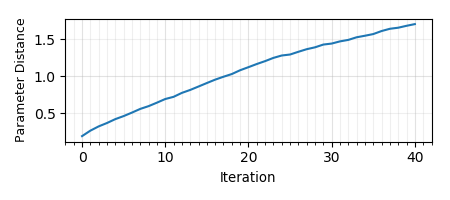}
\end{subfigure}
\caption{Plots for VGG-11 Epoch 1 trained using \textbf{SGD} on CIFAR-10. The descriptions of the plots are same as in Figure \ref{fig:loss_interpolation_gd_vgg11_cifar10_boom}. \textbf{Gist}: The loss interpolation between consecutive iterations have a minimum for iterations and cosine is less negative compared with GD, suggesting the optimization is oscillating along the walls of a valley like structure but doing more exploration compared with GD. This is verified by the larger distance traveled by SGD compared with GD in Figure \ref{fig:loss_interpolation_gd_vgg11_cifar10_boom}. The valley floor (dashed orange line) has many ups and downs showing barriers along SGD's path which do not affect its dynamics because SGD travels at a height above the floor. \label{fig:interpolation_vgg11_cifar10_epoch1}}
\vspace{-10pt}
\end{wrapfigure}
We now begin our analysis of studying the loss surface of DNNs along the trajectory of optimization updates. Specifically, consider that the parameters $\mathbf{\theta}$ of a DNN are initialized to a value $\mathbf{\theta}_0$. When using an optimization method to update these parameters, the $t^{th}$ update step takes the parameter from $\mathbf{\theta}_{t}$ to $\mathbf{\theta}_{t+1}$ using estimated gradient $\mathbf{g}_t$ as,
\begin{align}
\small
\mathbf{\theta}_{t+1} = \mathbf{\theta}_t - \eta \mathbf{g}_t
\end{align}
where $\eta$ is the learning rate. Notice the $t^{th}$ update step implies the $t^{th}$ epoch \textit{only} in the case when using the full batch gradient descent (GD-- gradient computed using the whole dataset). In the case of stochastic gradient descent, one iteration is an update from gradient computed from a mini-batch. We then interpolate the DNN loss between the convex combination of  $\mathbf{\theta}_{t}$ and $\mathbf{\theta}_{t+1}$ by considering parameter vectors $\mathbf{\theta}_{t}^\alpha = (1-\alpha) \theta_t + \alpha \theta_{t+1}$, where $\alpha \in [0,1]$ is chosen such that we obtain $10$ samples uniformly placed between these two parameter points. Simultaneously, we also keep track of two metrics-- the cosine of the angle between two consecutive gradients $\cos(\mathbf{g}_{t-1}, \mathbf{g}_t): = \mathbf{g}_{t-1}^T \mathbf{g}_{t}/(\lVert \mathbf{g}_{t-1}\rVert_2 \lVert \mathbf{g}_{t}\rVert_2)$, and the distance of the current parameter $\mathbf{\theta}_{t}$ from the initialization $\mathbf{\theta}_{0}$ given by $\lVert \theta_t - \theta_0 \rVert_2$. As it will become apparent in a bit, these two metrics along with the interpolation curve help us make deductions about how the optimization \textit{interacts} with the loss surface during its trajectory.

We perform experiments on MNIST \cite{mnistlecun}, CIFAR-10 \cite{cifar} and a subset of the tiny Imagenet dataset \cite{ILSVRC15} using multi-layer perceptrons (MLP), VGG-11 \cite{simonyan2014very} and Resnet-56 \cite{he2016deep} architectures with various batch sizes and learning rates. 
In the main text, we mostly show results for CIFAR-10 using VGG-11 architecture with a batch size of 100 and a fixed learning rate of 0.1 due to space limitations. Experiments on all the other datasets, architectures and hyper-parameter settings can be found in the appendix. All the claims are consistent across them.

\subsection{Optimization Trajectory}

We first experiment with full batch gradient descent (GD) to study its behavior before jumping to the analysis of SGD to isolate the confounding factor of mini-batch induced stochasticity. The plot of training loss interpolation between consecutive iterations (referred in the figure as \textit{training loss}), $\cos(\mathbf{g}_{t-1}, \mathbf{g}_t)$, and \textit{parameter distance} $\lVert \theta_t - \theta_0 \rVert_2$ for CIFAR-10 on VGG-11 architecture optimized using full batch gradient descent is shown in Figure \ref{fig:loss_interpolation_gd_vgg11_cifar10_boom} for the first $40$ iterations of training. To be clear, the x-axis is calibrated by the number of iterations, and there are $10$ interpolated loss values between each consecutive iterations (vertical gray lines) in the \textit{training loss} plot which is as described above (the cosine and parameter distance plots do not have any interpolations). This figure shows that the interpolated loss between every consecutive parameters from GD optimization update after iteration $15$ appears to be a quadratic-like structure with a minimum in between. 

Additionally, the cosine of the angle  between consecutive gradients after iteration $15$ is going negative and finally very close to $-1$, which means the consecutive gradients are almost along opposite directions. These two observations together suggest that the GD iterate is bouncing between walls of a valley-like landscape. For the iterations where there is a minimum in the interpolation between two iterations, we refer to this minimum as the \textit{floor} of the valley (these valley floors are connected by dashed orange line in figure \ref{fig:loss_interpolation_gd_vgg11_cifar10_boom} for clarity). We will see GD behavior shows lack of exploration for better minima by comparing its parameter distance from initialization with that of SGD. Take note that the parameter distance of GD during these $40$ iterations reaches $\sim 1.4$.

Now we perform the same analysis for SGD. Notice that even though the updates are performed using mini-batches for SGD, the training loss values used in the plot are computed using the full dataset to visualize the actual loss landscape. We show these plots for epoch $1$ (Figure \ref{fig:interpolation_vgg11_cifar10_epoch1}) in the main text and epoch $25$ (Figure \ref{fig:interpolation_vgg_epoch25}) and epoch $100$ (Figure \ref{fig:interpolation_vgg_epoch100}) in appendix. We find that while the loss interpolation also shows a quadratic-like structure with a minimum in between (similar to GD), there are some qualitative differences compared with GD. We see that the cosine of the angle between gradients from consecutive iterations are significantly less negative, suggesting that instead of oscillating in the same region, SGD is quickly moving away from its previous position. This can be verified by the parameter distance from initialization. We see that the distance after $40$ iterations is $\sim 1.7$, which is larger than the distance moved by GD\footnote{In general, after the same number of updates, GD traverses a smaller distance compared with SGD, see \cite{hoffer2017train}}. Finally and most interestingly, we see that the height of valley floor has many ups and downs for consecutive iterations in contrast with that of GD (emphasized by the dashed orange line in figure \ref{fig:interpolation_vgg11_cifar10_epoch1}), which means that there is a rough terrain or barriers along the path of SGD that could hinder exploration if the optimization was traveling too close to the valley floor. 

\begin{wrapfigure}{l}{0.42\textwidth}
\vspace{-10pt}
\begin{tabular}{|c|c|c|c|c|}
      \hline
     \textbf{Arch\textbackslash Epochs} &\textbf{ 1} & \textbf{ 10} & \textbf{ 25}& \textbf{ 100}\\\hline
      \textbf{VGG-11}&0 & 0 & 5 &13\\\hline
      \textbf{Resnet-56}& 0& 0& 2 &23\\\hline
      \textbf{MLP}&0&3&5&-\\\hline
    \end{tabular}
    \captionof{table}{\small Number of barriers crossed during training of one epoch (450 iterations) for VGG-11 and Resnet-56 on CIFAR-10 and MLP on MNIST. We say a barrier is crossed during an update step if there exists a point interpolated between the parameters before and after an update which has a loss value higher than the loss at either points. For most parts of the training, we find that SGD does not cross any significant number of barriers.\label{tab:barrier}}
\vspace{-15pt}
\end{wrapfigure}
A similar analysis for Resnet-56 on CIFAR-10, MLP on MNIST, VGG-11 on tiny ImageNet trained using GD for the first epoch are shown in Figures \ref{fig:interpolation_resnet_gd_boom}, \ref{fig:interpolation_MLP_gd},\ref{fig:interpolation_imagnet_gd} respectively in appendix. The same experiments analysis for SGD on different datasets, architectures under different hyper-parameters are also shown in section 1 in appendix. The observations and rules we discovered and described here are all consistent for all these experiments. 

In order to show that the claim about optimization not crossing barriers extends to the whole training instead of only a few iterations we've shown, we quantitatively measure for the entire epoch in different phase of training  if barriers are crossed. This result is shown in table \ref{tab:barrier} for VGG-11 and Resnet-56 trained on CIFAR-10 (trained for 100 epochs) and an MLP trained on MNIST (trained for 40 epochs). As we see, no barriers are crossed for most parts of the training. We further compute the number of barriers crossed for the first $40$ epochs for VGG-11 on CIFAR-10 shown in Figure \ref{fig:barrier} in the appendix: no barriers are crossed for most of the epochs and even for the barriers that are crossed towards the end, we find that their heights are substantially smaller compared with the loss value at the corresponding point during training, meaning they are not significant.

\begin{wrapfigure}{r}{0.42\textwidth}
\vspace{-15pt}
 \includegraphics[scale=0.5,trim=0.in 0.1in 0.in 0.in,clip]{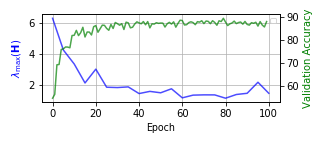}
\captionof{figure}{Spectral norm of the Hessian and validation accuracy for VGG-11 trained on CIFAR-10 using a fixed learning rate of 0.1 and batch size 100. Spectral norm and validation accuracy are roughly inversely correlated.}
\label{fig:vgg11_cifar10_hessian_valacc}
\vspace{-10pt}
\end{wrapfigure}

Finally, we track the spectral norm of the Hessian along with the validation accuracy while the model is being trained\footnote{Note that we track the spectral norm in the train mode of batchnorm; we observe that in validation mode the values are significantly larger. Tracking the value in train mode is fair because this is what SGD \textit{experiences} during training. Additionally, we track spectral norm because it captures the largest eigenvalue of the Hessian in contrast with Frobenius norm which can be misleading because the Hessian may have negative eigenvalues and the Frobenius norm sums the square of all eigenvalues.}. This plot is shown in Figure \ref{fig:vgg11_cifar10_hessian_valacc} for VGG-11 (and Figure \ref{fig:resnet56_cifar10_hessian_valacc} for Resnet-56 in the appendix). We find that the spectral norm reduces as training progresses (hence SGD finds flatter regions) but starts increasing towards the end. This is mildly correlated with a drop in validation accuracy towards the end. Regarding this correlation, while \citet{Dinh-et-al-2017} discuss that sharper minima can perform as well as wider ones, it is empirically known that flatter minima generalize better than sharper ones with SGD \citep{keskar2016large,jastrzkebski2017three}. This may be explained by \citet{neyshabur2017exploring, achille2017emergence} that discuss that minima that are both wide and have small norm may explain generalization in over-parametrized deep networks.
%
\vspace{-10pt}
\subsection{Qualitative Roles of Learning Rate and Batch Size}
\begin{wrapfigure}{r}{0.48\textwidth}
 \vspace{-30pt}
 \centering
\begin{subfigure}[c]{0.45\columnwidth}
  \centering
  \includegraphics[scale=0.6,trim=0.18in 0.35in 0.in 0.in,clip]{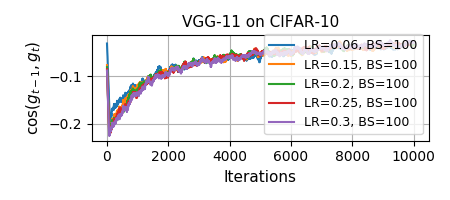}
\end{subfigure}
\vfill
\begin{subfigure}[c]{0.45\columnwidth}
  \centering
  \includegraphics[scale=0.6,trim=0.18in 0.1in 0.in 0.in,clip]{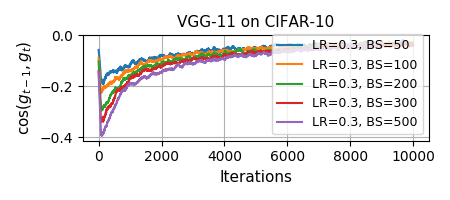}
\end{subfigure}
\caption{Changing batch size changes the cosine of angle between consecutive gradients while changing learning rate does not have any significant effect on the cosine. This shows batch size has a qualitatively different role compared with learning rate. Note that the curves are smoothened for visual clarity.\label{fig:cos_vgg11_cifar10}}
 \vspace{-15pt}
\end{wrapfigure}

We now focus in more detail on how the learning rate and batch size play qualitatively different roles during SGD optimization. As an extreme case, we already saw in the last section that when using GD vs SGD, the cosine of the angle between gradients from two consecutive iterations $\cos(\mathbf{g}_{t-1}, \mathbf{g}_t)$ is significantly closer to -1 (180 degrees) in the case of GD in contrast with SGD. 
Now we show on a more granular scale that changing the batch size gradually (keeping the learning rate fixed) changes $\cos(\mathbf{g}_{t-1}, \mathbf{g}_t)$, while changing the learning rate gradually (keeping batch size fixed) does not.
It is shown in Figure \ref{fig:cos_vgg11_cifar10} for VGG-11 trained on CIFAR-10 and in Figures \ref{fig:cos_resnet_cifar10}, \ref{fig:cos_mlp_mnist} and \ref{fig:cos_vgg_tiny}for  Resnet-56 with CIFAR-10, MLP with MNIST and VGG-11 with tiny ImageNet separately in the appendix. Notice the cosine is significantly more negative for larger batch sizes, implying that for larger batch sizes, the optimization is bouncing more within the same region instead of traveling farther along the valley as the case of small batch sizes.
This behavior is verified by the smaller distance of parameters from initialization during training for larger batch size which is also discussed by \citet{hoffer2017train}. This suggests that the noise from a small mini-batch size facilitates exploration that may lead to better minima and that this is hard to achieve by changing the learning rate.

On the other hand, we find that the learning rate controls the height from the valley floor at which the optimization oscillates along the valley walls which is important for avoiding barriers along SGD's path. Specifically, to quantify the height at which the optimization is bouncing above the valley floor, we make the following computations. Suppose at iterations $t$ and $t+1$ of training, the parameters are given by $\mathbf{\theta}_t$ and $\mathbf{\theta}_{t+1}$ respectively, and from the $10$ points sampled uniformly between $\mathbf{\theta}_t$ and $\mathbf{\theta}_{t+1}$ given by $\mathbf{\theta}_t^{\alpha}  = (1-\alpha) \theta_t + \alpha \theta_{t+1}$ for different values of $\alpha \in [0,1]$, we define $\mathbf{\theta}_t^{\min} := \mathbf{\theta}_t^{\arg \min_{\alpha} \mathcal{L}(\mathbf{\theta}_t^{\alpha})}$, where $\mathcal{L}(\mathbf{\theta})$ denotes the DNN loss at parameter $\mathbf{\theta}$ using the whole training set. Then we define the height of the iterate from the valley floor at iteration $t$ as $\frac{\mathcal{L}(\mathbf{\theta}_t) + \mathcal{L}(\mathbf{\theta}_{t+1}) - 2\mathcal{L}(\mathbf{\theta}_t^{\min})}{2}$. We then separately compute the average height for all iterations of epochs $1$, $10$, $25$ and $100$. 
\begin{wrapfigure}{l}{0.65\textwidth}
\vspace{-4pt}
\small
  \begin{center}
    \begin{tabular}{|c|c|c|c|c|} 
      \hline
      VGG-11&\textbf{Epoch 1} & \textbf{Epoch 10} & \textbf{Epoch 25}\\
      \hline
      \textbf{LR 0.1}&0.0625$\pm$1.2e-3 & 0.0199$\pm$4.8e-4 & 0.0104$\pm$2.2e-5\\\hline
      \textbf{LR 0.05}&0.0102$\pm$3.8e-5 & 0.0050$\pm$2.8e-5& 0.0035$\pm$1.7e-5 \\\hline
            \hline
      Resnet-56&\textbf{Epoch 1} & \textbf{Epoch 10} & \textbf{Epoch 25}\\
      \hline
      \textbf{LR 0.3}&0.0380$\pm$5.9e-4 & 0.0131$\pm$4.5e-4& 0.0094$\pm$1.3e-5\\\hline
      \textbf{LR 0.15}&0.0084$\pm$5.2e-5& 0.0034$\pm$3.2e-5& 0.0020$\pm$7.2e-6\\\hline
    \end{tabular}
    \captionof{table}{Average height of SGD above valley floor across the iterations in one epoch for different epochs during the training of VGG-11 and Resnet-56 using SGD on CIFAR-10. Here the height at iteration $t$ is defined defined by $\frac{\mathcal{L}(\mathbf{\theta}_t) + \mathcal{L}(\mathbf{\theta}_{t+1}) - 2\mathcal{L}(\mathbf{\theta}_t^{\min})}{2}$. Results of Epoch 100 and of VGG-11 on Tiny-ImageNet can be found at Table \ref{tab:height_ref} in the appendix. \label{tab:height_vgg}}
  \end{center}
\vspace{-15pt}
\end{wrapfigure}
These values are shown in table \ref{tab:height_vgg} for VGG-11 and Resnet-56 trained on CIFAR-10 and in table \ref{tab:height_ref} in appendix for VGG-11 on Tiny-ImageNet. They show that for almost all epochs, a smaller learning rate leads to a smaller height from the valley floor. Since the floor has barriers, it would increase the risk of hindering exploration for flatter minima. This has been corroborated by the recent empirical observations that smaller learning rates lead to sharper minima and poor generalization \cite{smith2017don, jastrzkebski2017three}. Based on our observations on the role of learning rate and batch-size, we empirically study learning rate schedules in appendix \ref{section_lr_sch}.

\vspace{-8pt}
\section{Importance of SGD Noise Structure}
\vspace{-5pt}
The gradient $\mathbf{g}_{SGD}(\mathbf{\theta})$ from mini-batch SGD at a parameter value $\mathbf{\theta}$ is expressed as,
$\mathbf{g}_{SGD}(\mathbf{\theta}) = \mathbf{\bar{g}}(\mathbf{\theta}) + \frac{1}{\sqrt{B}} \mathbf{n} (\mathbf{\theta})$,
where $\mathbf{n}(\mathbf{\theta})\sim \mathcal{N}(0, \mathbf{C}(\mathbf{\theta}))$, $ \mathbf{\bar{g}}(\mathbf{\theta})$ denotes the expected gradient using all training samples, $B$ is the mini-batch size and $\mathbf{C}(\mathbf{\theta})$ is the gradient covariance matrix at $\mathbf{\theta}$. In the previous section we discussed how mini-batch induced stochasticity plays a crucial role in SGD based optimization. This stochasticity due to SGD has historically been attributed to helping the optimization escape local minima in DNNs. However, the importance of the structure of the gradient covariance matrix $\mathbf{C}(\mathbf{\theta})$ is often neglected in these claims. To better understand its importance, we study the training dynamics of full batch gradient with artificially added isotropic noise. Specifically, we treat isotropic noise as our null hypothesis to confirm that the structure of noise induced by mini-batches in SGD is important.

In this experiment, we first train our models with gradient descent (GD), meaning there is no noise sampled from $\mathcal{N}(0, \mathbf{C}(\mathbf{\theta}))$ of SGD. For gradient descent with isotropic noise, we add isotropic noise at every iteration on $\mathbf{\bar{g}}(\mathbf{\theta})$. The noise is sampled from a normal distribution with variance calculated by multiplying the maximum gradient variance of the model at the initialization with a factor of $0.1$ and $0.05$. We train all models until their training losses saturate and monitor training loss, validation accuracy, cosine of the angle between gradients from two consecutive iterations and the parameter distance from initialization. 

\begin{wrapfigure}{r}{0.48\textwidth}
  \centering
  \includegraphics[scale=0.5,trim=0.in 0.35in 0.in 0.in,clip]{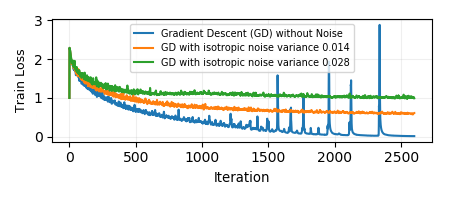}
  \includegraphics[scale=0.5,trim=0.in 0.35in 0.in 0.1in,clip]{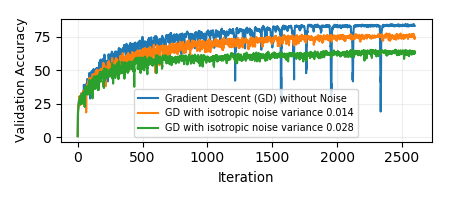}
  \includegraphics[scale=0.5,trim=0.17in 0.35in 0.in 0.in,clip]{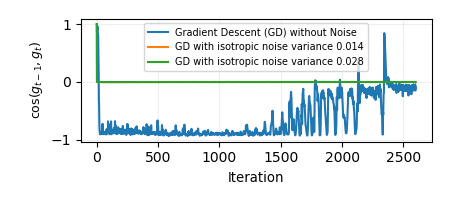}
  \includegraphics[scale=0.5,trim=0.in 0.1in 0.in 0.in,clip]{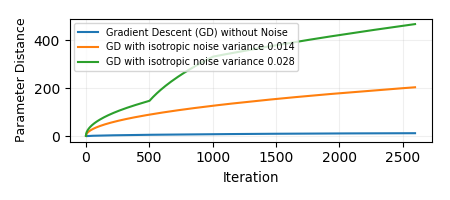}
\captionof{figure}{Plots for VGG-11 trained by GD (without noise) and GD with artificial isotropic noise sampled from Gaussian distribution with different variances. Models trained using GD with added isotropic noise get stuck in terms of training loss and have worse validation performance compared with the model trained with GD.}
\label{fig:vgg_isotropic}
\vspace{-35pt}
\end{wrapfigure}

 Figure \ref{fig:vgg_isotropic} shows the results for VGG-11 and Figures \ref{fig:resnet_isotropic} and  \ref{fig:mlp_isotropic} in the appendix shows the results for Resnet-56 on CIFAR-10 and MLP on MNIST. From the training loss and validation accuracy curves, we can see that adding even a small isotropic noise makes both the convergence\footnote{We additionally find that the model trained with isotropic noise gets stuck because we find that neither reducing learning rate, nor switching to GD at this point leads to reduction in training loss. However, switching to SGD makes the loss go down.} and generalization worse compared with the model trained with GD. The cosine of the angle between gradients of two consecutive iterations is close to $-1$ for $1500$ iterations for GD, which means two consecutive gradients are almost along opposite directions. It is an extra evidence that GD makes the optimization bounce off valley walls, which is what we discussed in section \ref{the_walk}. The parameter distance from initialization shows that models trained with isotropic noise travel farther away compared with the model trained using noiseless GD. These distances are much larger even compared with models trained with SGD (not shown here) for the same number of updates. 

To gain more intuitions into this behavior, we also calculate the norm of the final parameters found by GD, SGD and the isotropic noise cases. The parameter norms are $87$ and $82$ for GD and SGD respectively, and $369$ and $443$ for the $0.014$ and $0.028$ isotropic variance case. These numbers corroborate the generally discussed notion that SGD finds solutions with small $\ell^2$ norm \cite{zhang2016understanding} compared with GD, and the fact that isotropic noise solutions have much larger norms and get stuck suggests that isotropic noise both hinders optimization and is bad for generalization.

\citet{neelakantan2015adding} suggest adding isotropic noise to gradients and report performance improvement on a number of tasks. However, notice the crucial difference between our claim in this section and their setup is that they add isotropic noise on top of the noise due to the mini-batch induced stochasticity, while we add isotropic noise to the full dataset gradient (hence no noise is sampled from the gradient covariance matrix).

To gains insights into why the noise sampled from the gradient covariance matrix $\mathbf{C}(\mathbf{\theta})$ helps SGD, we note that there is a relationship between the covariance $\mathbf{C}(\mathbf{\theta})$ and the Hessian $\mathbf{H}(\mathbf{\theta})$ of the loss surface at parameter $\mathbf{\theta}$ which is revealed by the generalized Gauss Newton decomposition (see \citet{sagun2017empirical}) when using the cross-entropy (or negative log likelihood) loss. Let $p_i(\theta)$ denote the predicted probability output (of the correct class in the classification setting for instance) of a DNN parameterized by $\theta$ for the $i^{th}$ data sample (in total N samples). Then the negative log likelihood loss for the $i^{th}$ sample is given by, $\mathcal{L}_i(\theta) = -\log(p_i(\theta))$. \textit{The relation between the Hessian} $\mathbf{H}(\mathbf{\theta})$ \textit{and the gradient covariance} $\mathbf{C}(\theta)$ \textit{for negative log likelihood loss is},
\begin{align}
\small
\vspace{-15pt}
\mathbf{H}(\mathbf{\theta}) = \mathbf{C}(\theta)  + \mathbf{\bar{g}}(\theta) \mathbf{\bar{g}}(\theta)^T +  \frac{1}{N} \sum_{i=1}^{N} \frac{\partial \mathcal{L}_i(\theta)}{\partial p_i(\theta)} \cdot \frac{\partial^2 p_i(\theta)}{\partial \theta^2}\nonumber
\vspace{-15pt}
\end{align}
The derivation can be found in section \ref{appendix_hessian_covariance} of the appendix.
Thus we find that the Hessian and covariance at any point $\theta$ are related, and are almost equal near minima where the second term tends to zero. This relationship would imply that the mini-batch induced noise is roughly aligned with sharper directions of the loss landscape (empirically confirmed concurrently by \cite{zhu2018regularization}). This would prevent the optimization from converging along such directions unless a wider region is found, which could explain why SGD finds wider minima without relying on the stochastic differential equation framework (previous work) which assumes a reasonably small learning rate.
\vspace{-8pt}
\section{Discussion}
\vspace{-8pt}
We presented qualitative results to understand how GD and SGD interact with the DNN loss surface and avoided assumptions in order to rely instead on empirical evidence. We now draw similarities between the optimization trajectory in DNNs that we have empirically found, with those in quadratic loss optimization (see section 5 of \citet{lecun1998efficient}). Based on our empirical evidence, we deduce that both GD and SGD move in a valley like landscape by bouncing off valley walls. This is reminiscent of optimization in a quadratic loss setting with a non-isotropic positive semi-definite Hessian, where the optimal learning rate $\eta$ causes under-damping without divergence along eigenvectors of the Hessian which have eigenvalues $\lambda_i$ such that $ \lambda_i^{-1}< \eta < 2\lambda_i^{-1}$. On the other hand, in the case of DNNs trained with GD, we find that even though the training loss oscillates between valley walls during consecutive iterations, the valley floor decreases smoothly (see Figure \ref{fig:loss_interpolation_gd_vgg11_cifar10_boom}). This is similar to the quadratic loss optimization with over-damped convergence along the eigenvectors corresponding to eigenvalues $\lambda_i$ such that $ \eta < \lambda_i^{-1}$. 

On a different note, it is commonly conjectured that when training DNNs, SGD crosses barriers to escape local minima. Contrary to this commonly held intuition, we find that SGD almost never crosses any significant barriers along its path. More interestingly, when training with a certain learning rate (see figure \ref{fig:interpolation_vgg11_cifar10_epoch1}), we find barriers at the floor of the valley but SGD avoids them by traveling at a height above the floor (due to large learning rate). Hence if we use a small learning rate, SGD should encounter such barriers and likely cross them. But we found in our experiments that this was not the case. This suggests that while in theory SGD is capable of crossing barriers (due to stochasticity), it does not do so because probably there exist other directions in such regions along which SGD can continue to optimize without crossing barriers. But since small learning rates empirically correlate with bad generalization, this suggests that moving over such barriers instead of crossing them by using a large learning rate is a good mechanism for exploration for good regions.

Finally, much of what we have discussed is based on the loss landscape of specific datasets and architectures along with network parameterization choices like rectified linear activation units (ReLUs) and batch normalization \cite{ioffe2015batch}. These conclusions may differ depending on these choices. In these cases analysis similar to ours can be performed to see if similar dynamics hold or not. Studying these dynamics may provide more practical guidelines for setting optimization hyperparameters.


\bibliography{references}
\bibliographystyle{plainnat}

\newpage
\input{appendix}

\end{document}

%% file: appendix.tex
\setcounter{page}{1}
\appendix
\section*{Appendix} 
\section{Optimization Trajectory}
\begin{figure}
 \centering
\begin{subfigure}[t]{1\columnwidth}
  \centering
  \includegraphics[width=\columnwidth ,trim=0.in 0.1in 0.in 0.in,clip]{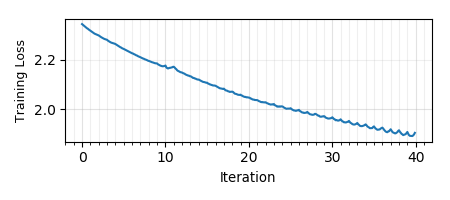}
\end{subfigure}\hspace{0.2\textwidth}
\begin{subfigure}[t]{1\columnwidth}
  \centering
  \includegraphics[width=\columnwidth ,trim=0.in 0.3in 0.in 0.in,clip]{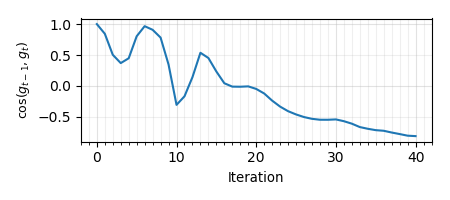}
\end{subfigure}\hspace{0.1\textwidth}
\begin{subfigure}[t]{1\columnwidth}
  \centering
  \includegraphics[width=\columnwidth ,trim=0.in 0.3in 0.in 0.in,clip]{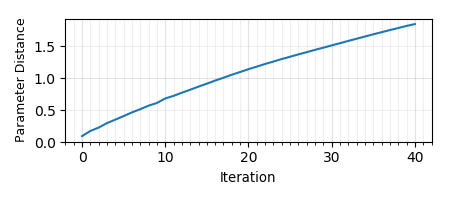}
\end{subfigure}\hspace{0.1\textwidth}
\caption{Plots for Resnet-56 Epoch 1 trained using full batch \textbf{Gradient Descent (GD)} on CIFAR-10. The descriptions of the plots are same as in figure \ref{fig:loss_interpolation_gd_vgg11_cifar10_boom}.}
\label{fig:interpolation_resnet_gd_boom}
\end{figure}

\begin{figure}
 \centering
\begin{subfigure}[t]{1\columnwidth}
  \centering
  \includegraphics[width=\columnwidth ,trim=0.in 0.3in 0.in 0.in,clip]{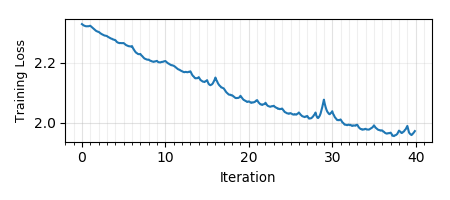}
\end{subfigure}\hspace{0.2\textwidth}
\begin{subfigure}[t]{1\columnwidth}
  \centering
  \includegraphics[width=\columnwidth ,trim=0.in 0.3in 0.in 0.in,clip]{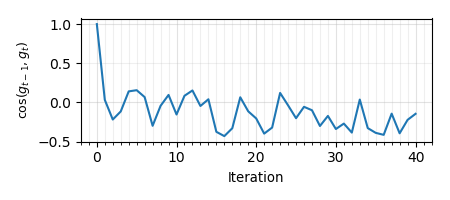}
\end{subfigure}\hspace{0.1\textwidth}
\begin{subfigure}[t]{1\columnwidth}
  \centering
  \includegraphics[width=\columnwidth ,trim=0.in 0.3in 0.in 0.in,clip]{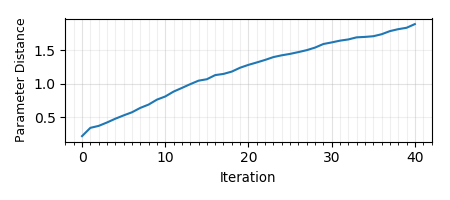}
\end{subfigure}\hspace{0.1\textwidth}
\caption{Plots for Resnet-56 Epoch 1 trained using \textbf{SGD} on CIFAR-10. The descriptions of the plots are same as in figure \ref{fig:loss_interpolation_gd_vgg11_cifar10_boom}.}
\label{fig:interpolation_resnet_epoch1}
\end{figure}

\begin{figure}
 \centering
\begin{subfigure}[t]{1\columnwidth}
  \centering
  \includegraphics[width=\columnwidth ,trim=0.in 0.3in 0.in 0.in,clip]{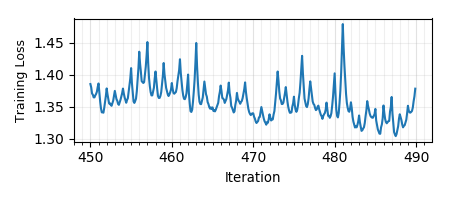}
\end{subfigure}\hspace{0.2\textwidth}
\begin{subfigure}[t]{1\columnwidth}
  \centering
  \includegraphics[width=\columnwidth ,trim=0.in 0.3in 0.in 0.in,clip]{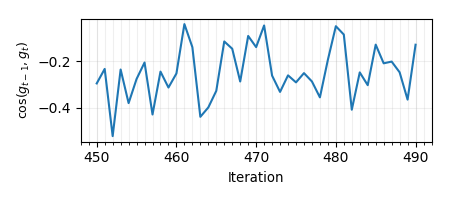}
\end{subfigure}\hspace{0.1\textwidth}
\begin{subfigure}[t]{1\columnwidth}
  \centering
  \includegraphics[width=\columnwidth ,trim=0.in 0.1in 0.in 0.in,clip]{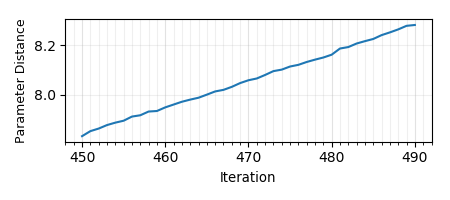}
\end{subfigure}\hspace{0.1\textwidth}
\caption{Plots for Resnet-56 Epoch 2 trained using \textbf{SGD} on CIFAR-10. The descriptions of the plots are same as in figure \ref{fig:loss_interpolation_gd_vgg11_cifar10_boom}.}
\label{fig:interpolation_resnet_epoch2}
\end{figure}

\begin{figure}
 \centering
\begin{subfigure}[t]{1\columnwidth}
  \centering
  \includegraphics[width=\columnwidth ,trim=0.in 0.3in 0.in 0.in,clip]{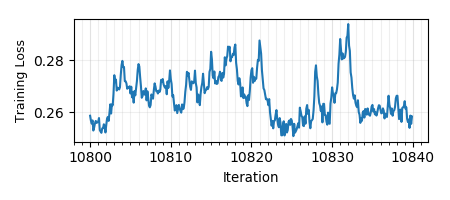}
\end{subfigure}\hspace{0.2\textwidth}
\begin{subfigure}[t]{1\columnwidth}
  \centering
  \includegraphics[width=\columnwidth ,trim=0.in 0.3in 0.in 0.in,clip]{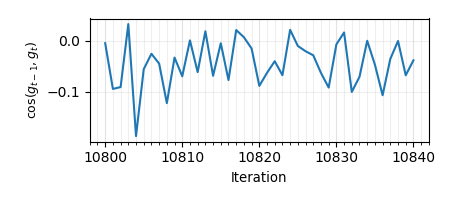}
\end{subfigure}\hspace{0.1\textwidth}
\begin{subfigure}[t]{1\columnwidth}
  \centering
  \includegraphics[width=\columnwidth ,trim=0.in 0.1in 0.in 0.in,clip]{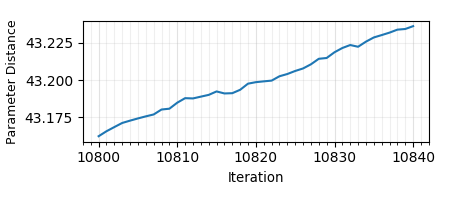}
\end{subfigure}\hspace{0.1\textwidth}
\caption{Plots for Resnet-56 Epoch 25 trained using \textbf{SGD} on CIFAR-10. The descriptions of the plots are same as in figure \ref{fig:loss_interpolation_gd_vgg11_cifar10_boom}.}
\label{fig:interpolation_resnet_epoch25}
\end{figure}

\begin{figure}
 \centering
\begin{subfigure}[t]{1\columnwidth}
  \centering
  \includegraphics[width=\columnwidth ,trim=0.in 0.3in 0.in 0.in,clip]{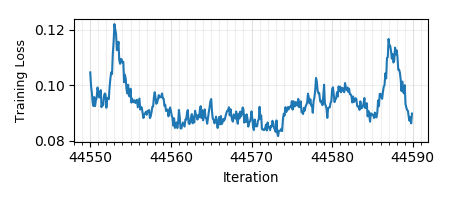}
\end{subfigure}\hspace{0.2\textwidth}
\begin{subfigure}[t]{1\columnwidth}
  \centering
  \includegraphics[width=\columnwidth ,trim=0.in 0.3in 0.in 0.in,clip]{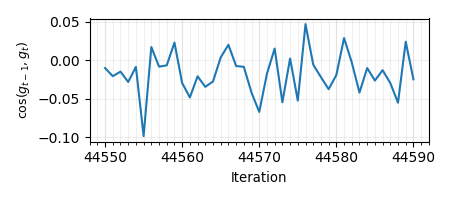}
\end{subfigure}\hspace{0.1\textwidth}
\begin{subfigure}[t]{1\columnwidth}
  \centering
  \includegraphics[width=\columnwidth ,trim=0.in 0.1in 0.in 0.in,clip]{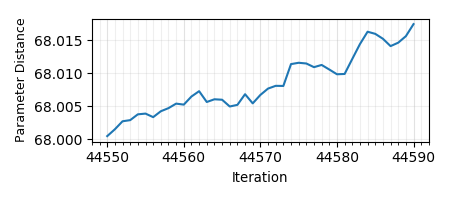}
\end{subfigure}\hspace{0.1\textwidth}
\caption{Plots for Resnet-56 Epoch 100 trained using \textbf{SGD} on CIFAR-10. The descriptions of the plots are same as in figure \ref{fig:loss_interpolation_gd_vgg11_cifar10_boom}.}
\label{fig:interpolation_resnet_epoch100}
\end{figure}

\begin{figure}
 \centering
\begin{subfigure}[t]{1\columnwidth}
  \centering
  \includegraphics[width=\columnwidth ,trim=0.in 0.3in 0.in 0.in,clip]{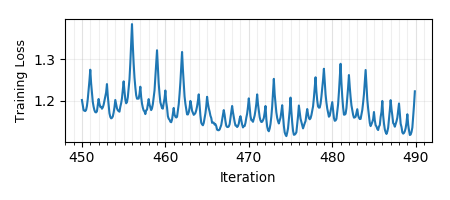}
\end{subfigure}\hspace{0.2\textwidth}
\begin{subfigure}[t]{1\columnwidth}
  \centering
  \includegraphics[width=\columnwidth ,trim=0.in 0.3in 0.in 0.in,clip]{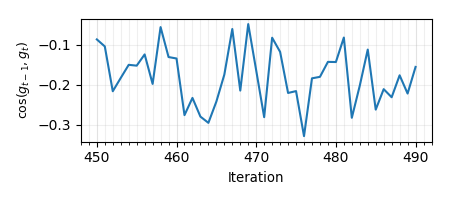}
\end{subfigure}\hspace{0.1\textwidth}
\begin{subfigure}[t]{1\columnwidth}
  \centering
  \includegraphics[width=\columnwidth ,trim=0.in 0.1in 0.in 0.in,clip]{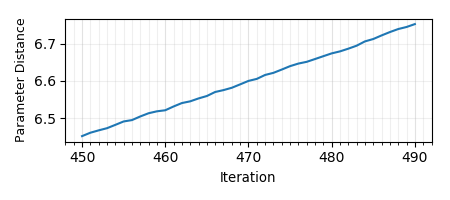}
\end{subfigure}\hspace{0.1\textwidth}
\caption{Plots for VGG-11 Epoch 2 trained using \textbf{SGD} on CIFAR-10. The descriptions of the plots are same as in figure \ref{fig:loss_interpolation_gd_vgg11_cifar10_boom}.}
\label{fig:interpolation_vgg_epoch2}
\end{figure}

\begin{figure}
 \centering
\begin{subfigure}[t]{1\columnwidth}
  \centering
  \includegraphics[width=\columnwidth ,trim=0.in 0.35in 0.in 0.in,clip]{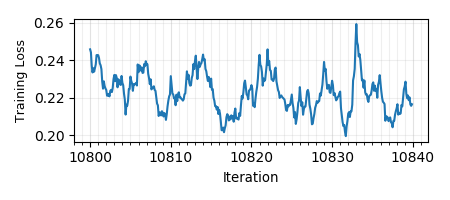}
\end{subfigure}\hspace{0.2\textwidth}
\begin{subfigure}[t]{1\columnwidth}
  \centering
  \includegraphics[width=\columnwidth ,trim=0.in 0.35in 0.in 0.15in,clip]{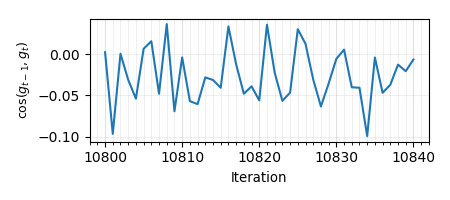}
\end{subfigure}\hspace{0.1\textwidth}
\begin{subfigure}[t]{1\columnwidth}
  \centering
  \includegraphics[width=\columnwidth ,trim=0.in 0.1in 0.in 0.15in,clip]{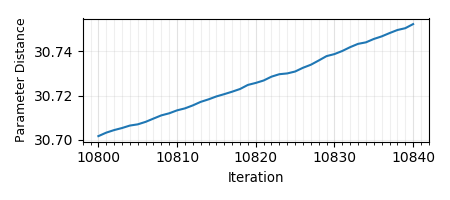}
\end{subfigure}\hspace{0.1\textwidth}
\caption{Plots for VGG-11 Epoch 25 trained using \textbf{SGD} on CIFAR-10. The descriptions of the plots are same as in figure \ref{fig:loss_interpolation_gd_vgg11_cifar10_boom}.}
\label{fig:interpolation_vgg_epoch25}
\end{figure}

\begin{figure}
 \centering
\begin{subfigure}[t]{1\columnwidth}
  \centering
  \includegraphics[width=\columnwidth ,trim=0.in 0.35in 0.in 0.in,clip]{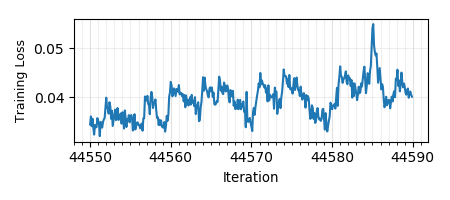}
\end{subfigure}\hspace{0.2\textwidth}
\begin{subfigure}[t]{1\columnwidth}
  \centering
  \includegraphics[width=\columnwidth ,trim=0.in 0.35in 0.in 0.15in,clip]{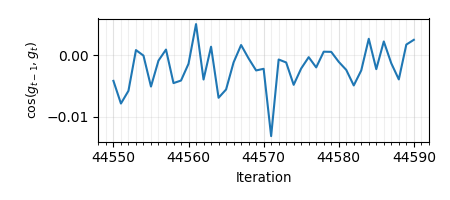}
\end{subfigure}\hspace{0.1\textwidth}
\begin{subfigure}[t]{1\columnwidth}
  \centering
  \includegraphics[width=\columnwidth ,trim=0.in 0.1in 0.in 0.15in,clip]{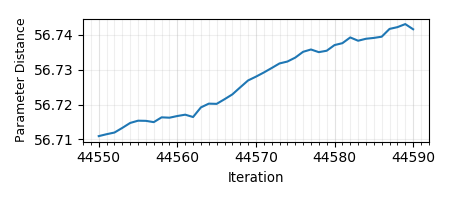}
\end{subfigure}\hspace{0.1\textwidth}
\caption{Plots for VGG-11 Epoch 100 trained using \textbf{SGD} on CIFAR-10. The descriptions of the plots are same as in figure \ref{fig:loss_interpolation_gd_vgg11_cifar10_boom}.}
\label{fig:interpolation_vgg_epoch100}
\end{figure}

\begin{figure}
 \centering
\begin{subfigure}[t]{1\columnwidth}
  \centering
  \includegraphics[width=\columnwidth ,trim=0.in 0.35in 0.in 0.in,clip]{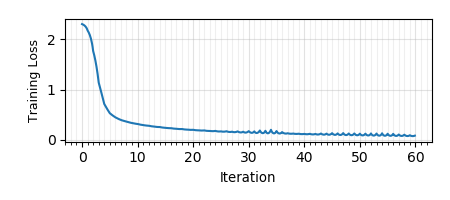}
\end{subfigure}\hspace{0.2\textwidth}
\begin{subfigure}[t]{1\columnwidth}
  \centering
  \includegraphics[width=\columnwidth ,trim=0.in 0.35in 0.in 0.15in,clip]{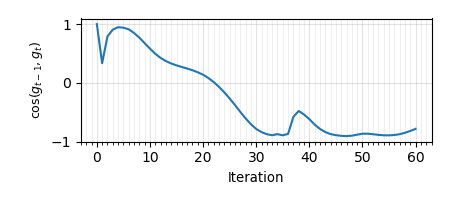}
\end{subfigure}\hspace{0.1\textwidth}
\begin{subfigure}[t]{1\columnwidth}
  \centering
  \includegraphics[width=\columnwidth ,trim=0.in 0.1in 0.in 0.15in,clip]{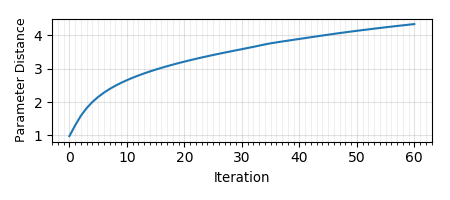}
\end{subfigure}\hspace{0.1\textwidth}
\caption{Plots for MLP Epoch 1 trained using full batch \textbf{Gradient Descent (GD)} on MNIST. The descriptions of the plots are same as in figure \ref{fig:loss_interpolation_gd_vgg11_cifar10_boom}.}
\label{fig:interpolation_MLP_gd}
\end{figure}

\begin{figure}
 \centering
\begin{subfigure}[t]{1\columnwidth}
  \centering
  \includegraphics[width=\columnwidth ,trim=0.in 0.3in 0.in 0.in,clip]{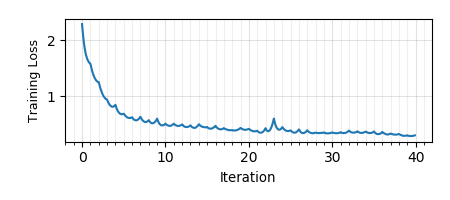}
\end{subfigure}\hspace{0.2\textwidth}
\begin{subfigure}[t]{1\columnwidth}
  \centering
  \includegraphics[width=\columnwidth ,trim=0.in 0.3in 0.in 0.in,clip]{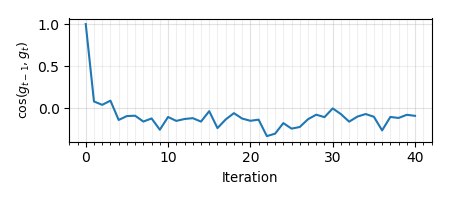}
\end{subfigure}\hspace{0.1\textwidth}
\begin{subfigure}[t]{1\columnwidth}
  \centering
  \includegraphics[width=\columnwidth ,trim=0.in 0.1in 0.in 0.in,clip]{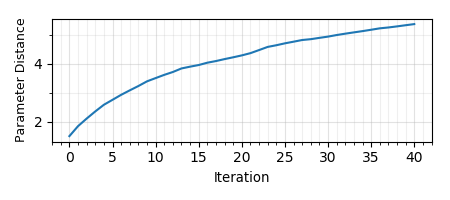}
\end{subfigure}\hspace{0.1\textwidth}
\caption{Plots for MLP Epoch 1 trained using \textbf{SGD} on MNIST. The descriptions of the plots are same as in figure \ref{fig:loss_interpolation_gd_vgg11_cifar10_boom}.}
\label{fig:interpolation_MLP_epoch1}
\end{figure}

\begin{figure}
 \centering
\begin{subfigure}[t]{1\columnwidth}
  \centering
  \includegraphics[width=\columnwidth ,trim=0.in 0.3in 0.in 0.in,clip]{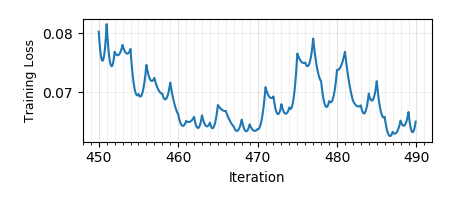}
\end{subfigure}\hspace{0.2\textwidth}
\begin{subfigure}[t]{1\columnwidth}
  \centering
  \includegraphics[width=\columnwidth ,trim=0.in 0.3in 0.in 0.in,clip]{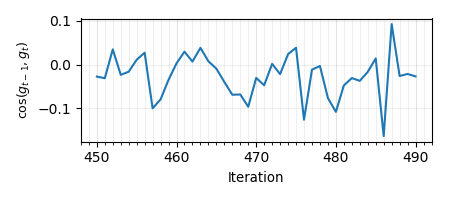}
\end{subfigure}\hspace{0.1\textwidth}
\begin{subfigure}[t]{1\columnwidth}
  \centering
  \includegraphics[width=\columnwidth ,trim=0.in 0.1in 0.in 0.in,clip]{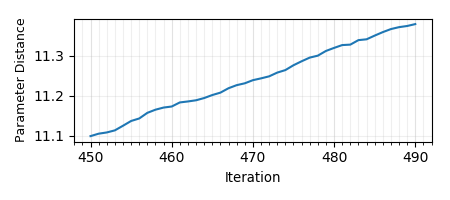}
\end{subfigure}\hspace{0.1\textwidth}
\caption{Plots for MLP Epoch 2 trained using \textbf{SGD} on MNIST. The descriptions of the plots are same as in figure \ref{fig:loss_interpolation_gd_vgg11_cifar10_boom}.}
\label{fig:interpolation_MLP_epoch2}
\end{figure}

\begin{figure}
 \centering
\begin{subfigure}[t]{1\columnwidth}
  \centering
  \includegraphics[width=\columnwidth ,trim=0.in 0.35in 0.in 0.in,clip]{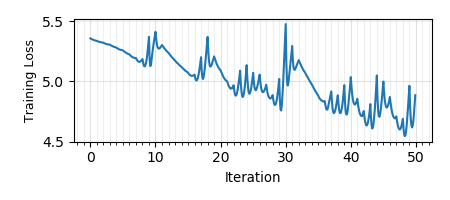}
\end{subfigure}\hspace{0.2\textwidth}
\begin{subfigure}[t]{1\columnwidth}
  \centering
  \includegraphics[width=\columnwidth ,trim=0.in 0.35in 0.in 0.15in,clip]{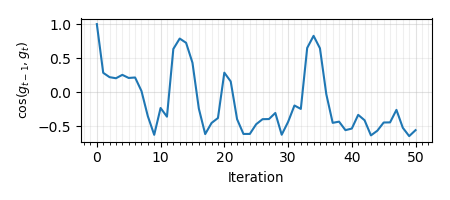}
\end{subfigure}\hspace{0.1\textwidth}
\begin{subfigure}[t]{1\columnwidth}
  \centering
  \includegraphics[width=\columnwidth ,trim=0.in 0.1in 0.in 0.15in,clip]{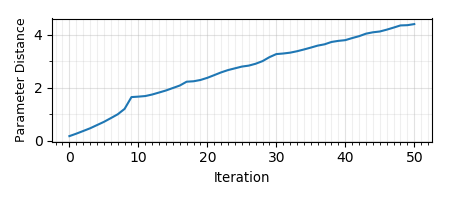}
\end{subfigure}\hspace{0.1\textwidth}
\caption{Plots for VGG-11 Epoch 1 trained using full batch \textbf{Gradient Descent (GD)} on Tiny-ImageNet. The descriptions of the plots are same as in figure \ref{fig:loss_interpolation_gd_vgg11_cifar10_boom}.}
\label{fig:interpolation_imagnet_gd}
\end{figure}

\begin{figure}
 \centering
\begin{subfigure}[t]{1\columnwidth}
  \centering
  \includegraphics[width=\columnwidth ,trim=0.in 0.35in 0.in 0.in,clip]{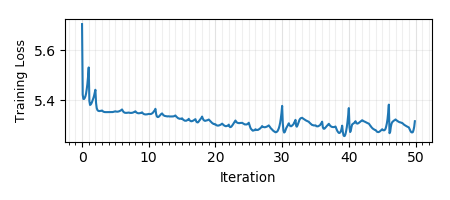}
\end{subfigure}\hspace{0.2\textwidth}
\begin{subfigure}[t]{1\columnwidth}
  \centering
  \includegraphics[width=\columnwidth ,trim=0.in 0.35in 0.in 0.15in,clip]{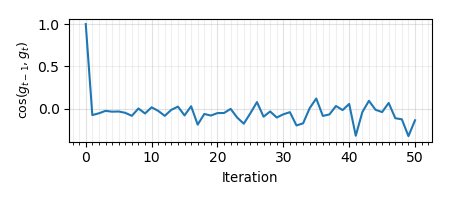}
\end{subfigure}\hspace{0.1\textwidth}
\begin{subfigure}[t]{1\columnwidth}
  \centering
  \includegraphics[width=\columnwidth ,trim=0.in 0.1in 0.in 0.15in,clip]{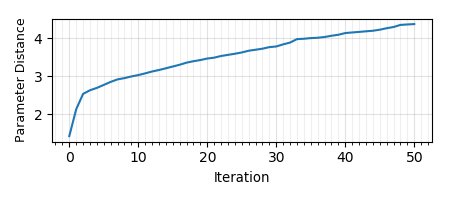}
\end{subfigure}\hspace{0.1\textwidth}
\caption{Plots for VGG-11 Epoch 1 trained using \textbf{SGD} on Tiny-ImageNet. The descriptions of the plots are same as in figure \ref{fig:loss_interpolation_gd_vgg11_cifar10_boom}.}
\label{fig:interpolation_imagnet_epoch1}
\end{figure}

\begin{figure}[!ht]
 \centering
\begin{subfigure}{1\columnwidth}
  \centering
  \includegraphics[scale=0.6,trim=0.in 0.35in 0.in 0.in,clip]{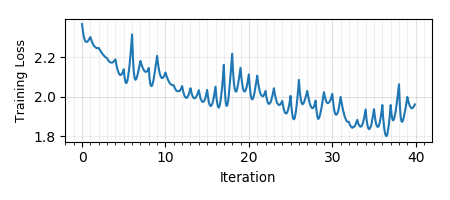}
\end{subfigure}\hspace{0.2\textwidth}
\begin{subfigure}[t]{1\columnwidth}
  \centering
    \includegraphics[scale=0.6,trim=0.in 0.35in 0.in 0.in,clip]{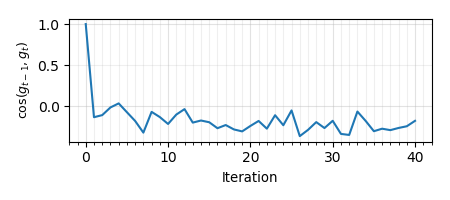}
\end{subfigure}\hspace{0.1\textwidth}
\begin{subfigure}{1\columnwidth}
  \centering
  \includegraphics[scale=0.6,trim=0.in 0.1in 0.in 0.in,clip]{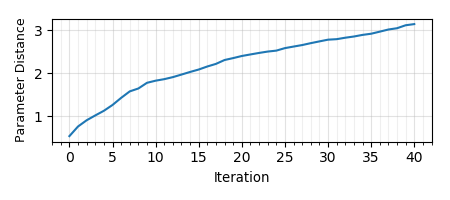}
\end{subfigure}
\caption{Plots for VGG-11 Epoch 1 trained using learning rate 0.3 batch size 100 on CIFAR-10.}
\label{fig:inter_vgg_0.3_100}
\end{figure}

\begin{figure}[!ht]
 \centering
\begin{subfigure}[t]{1\columnwidth}
  \centering
  \includegraphics[scale=0.6,trim=0.in 0.35in 0.in 0.in,clip]{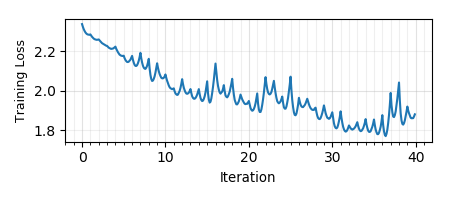}
\end{subfigure}\hspace{0.2\textwidth}
\begin{subfigure}[t]{1\columnwidth}
  \centering
   \includegraphics[scale=0.6,trim=0.in 0.35in 0.in 0.in,clip]{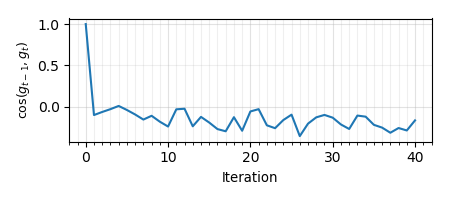}
\end{subfigure}\hspace{0.1\textwidth}
\begin{subfigure}[t]{1\columnwidth}
  \centering
  \includegraphics[scale=0.6,trim=0.in 0.1in 0.in 0.in,clip]{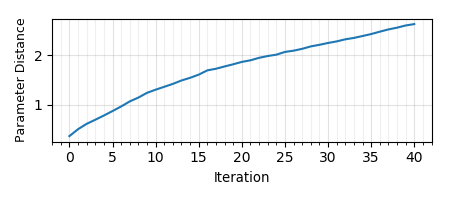}
\end{subfigure}
\caption{Plots for VGG-11 Epoch 1 trained using learning rate 0.2 batch size 100 on CIFAR-10.}
\label{fig:inter_vgg_0.2_100}
\end{figure}

\begin{figure}[!ht]
 \centering
\begin{subfigure}[t]{1\columnwidth}
  \centering
  \includegraphics[scale=0.6,trim=0.in 0.35in 0.in 0.in,clip]{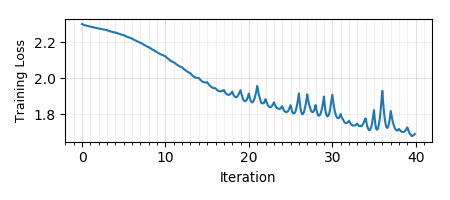}
\end{subfigure}\hspace{0.2\textwidth}
\begin{subfigure}[t]{1\columnwidth}
  \centering
   \includegraphics[scale=0.6,trim=0.in 0.35in 0.in 0.in,clip]{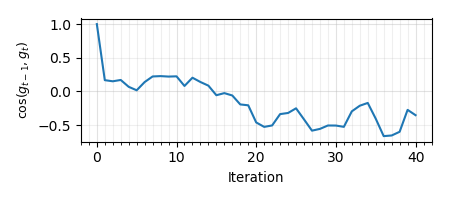}
\end{subfigure}\hspace{0.1\textwidth}
\begin{subfigure}[t]{1\columnwidth}
  \centering
  \includegraphics[scale=0.6,trim=0.in 0.1in 0.in 0.in,clip]{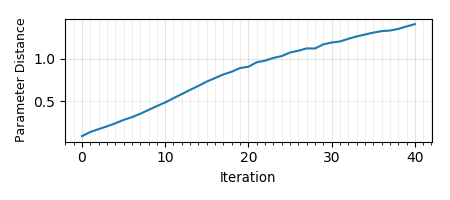}
\end{subfigure}
\caption{Plots for VGG-11 Epoch 1 trained using learning rate 0.1 batch size 500 on CIFAR-10.}
\label{fig:inter_vgg_0.1_500}
\end{figure}

\begin{figure}[!ht]
 \centering
\begin{subfigure}[t]{1\columnwidth}
  \centering
  \includegraphics[scale=0.6,trim=0.in 0.35in 0.in 0.in,clip]{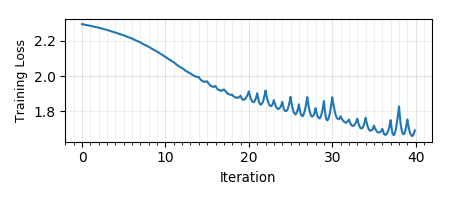}
\end{subfigure}\hspace{0.2\textwidth}
\begin{subfigure}[t]{1\columnwidth}
  \centering
   \includegraphics[scale=0.6,trim=0.in 0.35in 0.in 0.in,clip]{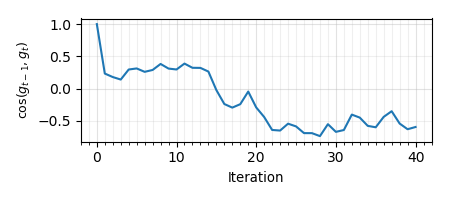}
\end{subfigure}\hspace{0.1\textwidth}
\begin{subfigure}[t]{1\columnwidth}
  \centering
  \includegraphics[scale=0.6,trim=0.in 0.1in 0.in 0.in,clip]{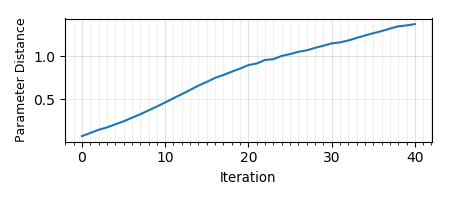}
\end{subfigure}
\caption{Plots for VGG-11 Epoch 1 trained using learning rate 0.1 batch size 1000 on CIFAR-10.}
\label{fig:inter_vgg_0.1_1000}
\end{figure}

\begin{figure}[!ht]
 \centering
\begin{subfigure}[t]{1\columnwidth}
  \centering
  \includegraphics[scale=0.6,trim=0.in 0.35in 0.in 0.in,clip]{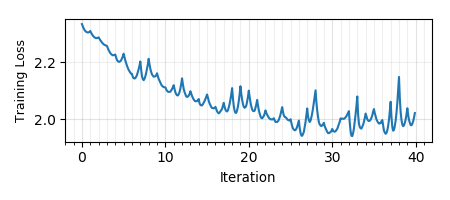}
\end{subfigure}\hspace{0.2\textwidth}
\begin{subfigure}[t]{1\columnwidth}
  \centering
   \includegraphics[scale=0.6,trim=0.in 0.35in 0.in 0.in,clip]{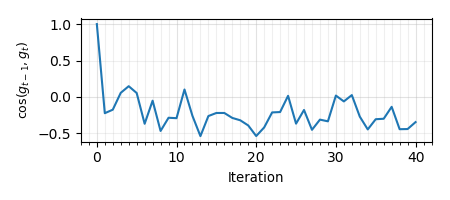}
\end{subfigure}\hspace{0.1\textwidth}
\begin{subfigure}[t]{1\columnwidth}
  \centering
  \includegraphics[scale=0.6,trim=0.in 0.1in 0.in 0.in,clip]{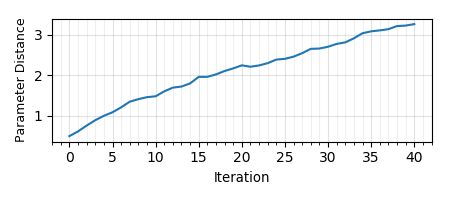}
\end{subfigure}
\caption{Plots for Resnet-56 Epoch 1 trained using learning rate 0.7 batch size 100 on CIFAR-10.}
\label{fig:inter_resnet_0.7_100}
\end{figure}

\begin{figure}[!ht]
 \centering
\begin{subfigure}[t]{1\columnwidth}
  \centering
  \includegraphics[scale=0.6,trim=0.in 0.35in 0.in 0.in,clip]{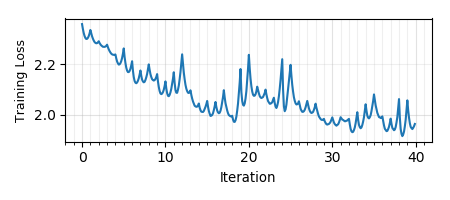}
\end{subfigure}\hspace{0.2\textwidth}
\begin{subfigure}[t]{1\columnwidth}
  \centering
   \includegraphics[scale=0.6,trim=0.in 0.35in 0.in 0.in,clip]{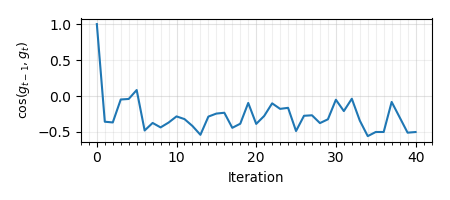}
\end{subfigure}\hspace{0.1\textwidth}
\begin{subfigure}[t]{1\columnwidth}
  \centering
  \includegraphics[scale=0.6,trim=0.in 0.1in 0.in 0.in,clip]{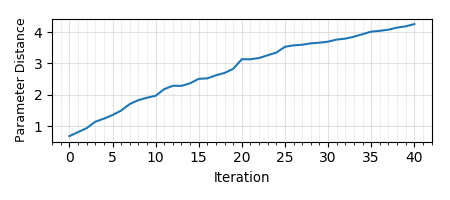}
\end{subfigure}
\caption{Plots for Resnet-56 Epoch 1 trained using learning rate 1 batch size 100 on CIFAR-10.}
\label{fig:inter_resnet_1_100}
\end{figure}

\begin{figure}[!ht]
 \centering
\begin{subfigure}[t]{1\columnwidth}
  \centering
  \includegraphics[scale=0.6,trim=0.in 0.35in 0.in 0.in,clip]{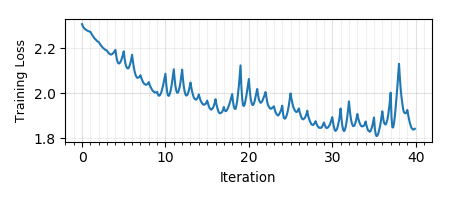}
\end{subfigure}\hspace{0.2\textwidth}
\begin{subfigure}[t]{1\columnwidth}
  \centering
   \includegraphics[scale=0.6,trim=0.in 0.35in 0.in 0.in,clip]{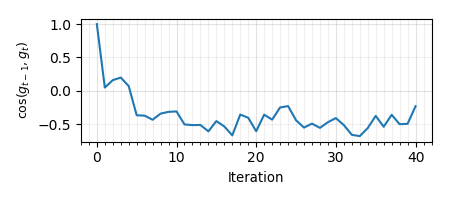}
\end{subfigure}\hspace{0.1\textwidth}
\begin{subfigure}[t]{1\columnwidth}
  \centering
  \includegraphics[scale=0.6,trim=0.in 0.1in 0.in 0.in,clip]{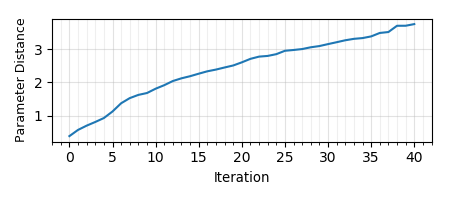}
\end{subfigure}
\caption{Plots for Resnet-56 Epoch 1 trained using learning rate 1 batch size 500 on CIFAR-10.}
\vspace{-15pt}
\label{fig:inter_resnet_1_500}
\end{figure}

\begin{figure}[!ht]
 \centering
\begin{subfigure}[t]{1\columnwidth}
  \centering
  \includegraphics[scale=0.6,trim=0.in 0.35in 0.in 0.in,clip]{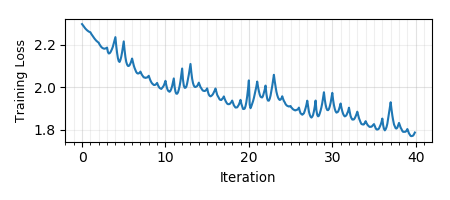}
\end{subfigure}\hspace{0.2\textwidth}
\begin{subfigure}[t]{1\columnwidth}
  \centering
   \includegraphics[scale=0.6,trim=0.in 0.35in 0.in 0.in,clip]{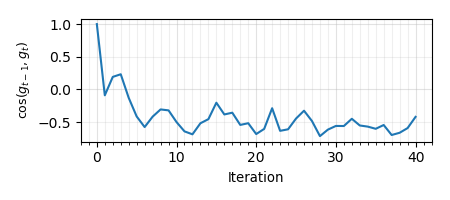}
\end{subfigure}\hspace{0.1\textwidth}
\begin{subfigure}[t]{1\columnwidth}
  \centering
  \includegraphics[scale=0.6,trim=0.in 0.1in 0.in 0.in,clip]{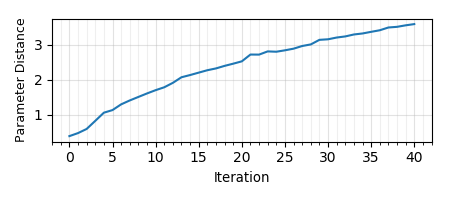}
\end{subfigure}
\caption{Plots for Resnet-56 Epoch 1 trained using learning rate 1 batch size 1000 on CIFAR-10.}
\vspace{-15pt}
\label{fig:inter_resnet_1_1000}
\end{figure}

This is a continuation of section $3.1$ in the main text. Here we show further experiments on other datasets, architectures and hyper-parameter settings. 
The analysis of GD training for Resnet-56 on CIFAR-10, MLP on MNIST and VGG-11 on tiny ImageNet are shown in figures \ref{fig:interpolation_resnet_gd_boom}, \ref{fig:interpolation_MLP_gd} and \ref{fig:interpolation_imagnet_gd} respectively. Similarly, the analysis of SGD training for Resnet-56 on CIFAR-10 dataset with batch size of 100 and learning rate 0.1 for epochs 1, 2, 25 and 100 are shown in figures \ref{fig:interpolation_resnet_epoch1}, \ref{fig:interpolation_resnet_epoch2}, \ref{fig:interpolation_resnet_epoch25} and \ref{fig:interpolation_resnet_epoch100} respectively. The analysis of SGD training for VGG-11 on CIFAR-10 with the batch size of 100 and learning rate 0.1 on epochs 2, 25,100 are shown in figures \ref{fig:interpolation_vgg_epoch2}, \ref{fig:interpolation_vgg_epoch25} and \ref{fig:interpolation_vgg_epoch100}. The analysis of SGD training for MLP on MNIST for epochs 1 and 2 are shown in figures \ref{fig:interpolation_MLP_epoch1} and \ref{fig:interpolation_MLP_epoch2}. The analysis of SGD training for VGG-11 on tiny ImageNet for epochs 1 is shown in figure \ref{fig:interpolation_imagnet_epoch1}. We also conducted the same experiment and analysis on various batch sizes and learning rates for every architecture. Results of VGG-11 can be found in figures \ref{fig:inter_vgg_0.3_100}, \ref{fig:inter_vgg_0.2_100}, \ref{fig:inter_vgg_0.1_500} and \ref{fig:inter_vgg_0.1_1000}. Results of Resnet-56 can be found in figures \ref{fig:inter_resnet_0.7_100}, \ref{fig:inter_resnet_1_100}, \ref{fig:inter_resnet_1_500} and \ref{fig:inter_resnet_1_1000}.  The observations and rules we discovered and described in section $3$ are all consistent for all these experiments. Specifically, for the interpolation of SGD for VGG-11 on tiny ImageNet, the valley-like trajectory is weird-looking but even so, according to our quantitative evaluation there is no barrier between any two consecutive iterations.

\begin{figure}[t]
\centering
 \includegraphics[scale=0.6,trim=0.in 0.1in 0.in 0.in,clip]{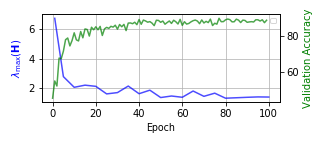}
\caption{Max eigenvalue (spectral norm) of the Hessian and validation accuracy for Resnet-56 trained on CIFAR-10 using a fixed learning rate of 0.3 and batch size 100. The spectral norm roughly decreases with training but starts increasing slightly towards the end. Similarly validation accuracy roughly improves throughout training but drops towards the end.}
\label{fig:resnet56_cifar10_hessian_valacc}
\end{figure}

We track the spectral norm of the Hessian along with the validation accuracy while the model is being trained. This is shown in figure \ref{fig:resnet56_cifar10_hessian_valacc} for Resnet-56 trained on CIFAR-10.

\section{Qualitative Roles of Learning Rate and Batch Size}
\begin{figure}[!ht]
 \centering
\begin{subfigure}[t]{1\columnwidth}
  \centering
  \includegraphics[scale=0.6,trim=0.in 0.35in 0.in 0.in,clip]{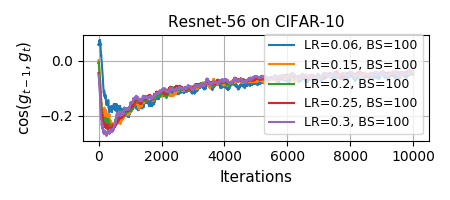}
\end{subfigure}
\vfill
\begin{subfigure}[t]{1\columnwidth}
  \centering
  \includegraphics[scale=0.6,trim=0.in 0.35in 0.in 0.in,clip]{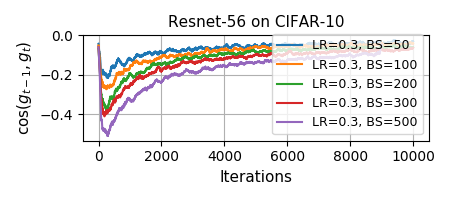}
\end{subfigure}
\caption{Changing batch size changes the cosine of angle between consecutive gradients while changing learning rate does not have any significant effect on the cosine. This shows batch size has a qualitatively different role compared with learning rate. Note that the curves are smoothened for visual clarity. \label{fig:cos_resnet_cifar10}}
\end{figure}

\begin{figure}[!ht]
 \centering
\begin{subfigure}[t]{1\columnwidth}
  \centering
  \includegraphics[scale=0.6,trim=0.in 0.35in 0.in 0.in,clip]{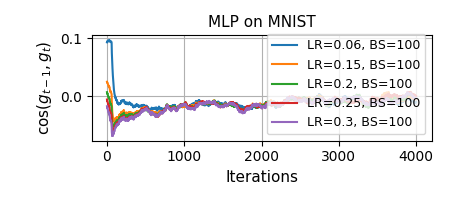}
\end{subfigure}
\vfill
\begin{subfigure}[t]{1\columnwidth}
  \centering
  \includegraphics[scale=0.6,trim=0.in 0.35in 0.in 0.in,clip]{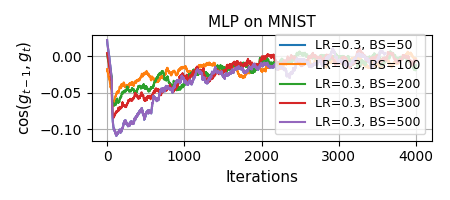}
\end{subfigure}
\caption{Changing batch size changes the cosine of angle between consecutive gradients while changing learning rate does not have any significant effect on the cosine. This shows batch size has a qualitatively different role compared with learning rate. Note that the curves are smoothened for visual clarity. \label{fig:cos_mlp_mnist}}
\end{figure}

\begin{figure}[!ht]
 \centering
\begin{subfigure}[t]{1\columnwidth}
  \centering
  \includegraphics[scale=0.6,trim=0.in 0.35in 0.in 0.in,clip]{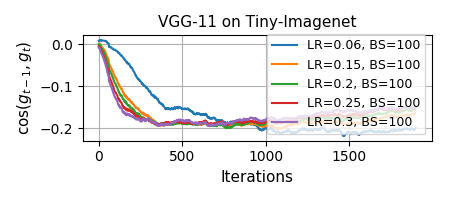}
\end{subfigure}
\vfill
\begin{subfigure}[t]{1\columnwidth}
  \centering
  \includegraphics[scale=0.6,trim=0.in 0.35in 0.in 0.in,clip]{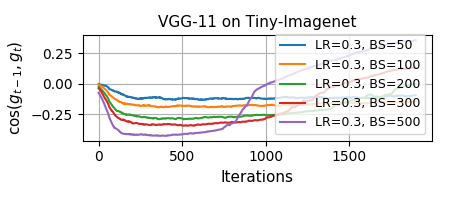}
\end{subfigure}
\caption{Changing batch size changes the cosine of angle between consecutive gradients while changing learning rate does not have any significant effect on the cosine. This shows batch size has a qualitatively different role compared with learning rate. Note that the curves are smoothened for visual clarity. \label{fig:cos_vgg_tiny}}
\end{figure}

This is a continuation of section $3.2$ in the main text. In this section we show further experiments for the analysis of different roles of learning rate and batch size  during training on various architectures and data sets. Figures \ref{fig:cos_resnet_cifar10}, \ref{fig:cos_mlp_mnist} and \ref{fig:cos_vgg_tiny} shows the results for Resnet-56 on CIFAR-10, MLP on MNIST and VGG-11 on Tiny-Imagenet. In all of the experiments, training the model with smaller batch size will make the angle between gradients of two consecutive iterations larger, which means for smaller batch size, instead of oscillating within the same region, the optimization travels farther along the valley, as we described in section $3.2$. For all architectures, changing learning rate doesn't change the angles. 
\begin{table*}
\small
  \begin{center}
    \caption{Average height of SGD above valley floor across the iterations in one epoch for different epochs during the training of VGG-11 and Resnet-56 using SGD on CIFAR-10 and VGG-11 on Tiny-ImageNet. Here the height at iteration $t$ is defined defined by $\frac{\mathcal{L}(\mathbf{\theta}_t) + \mathcal{L}(\mathbf{\theta}_{t+1}) - 2\mathcal{L}(\mathbf{\theta}_t^{\min})}{2}$.  \label{tab:height_ref}}
    \begin{tabular}{|c|c|c|c|c|} 
      \hline
      VGG-11 on CIFAR-10&\textbf{Epoch 1} & \textbf{Epoch 10} & \textbf{Epoch 25}& \textbf{Epoch 100}\\
      \hline
      \textbf{LR 0.1}&0.0625$\pm$1.2e-3 & 0.0199$\pm$4.8e-4 & 0.0104$\pm$2.2e-5&0.0025$\pm$3.1e-6\\\hline
      \textbf{LR 0.05}&0.0102$\pm$3.8e-5 & 0.0050$\pm$2.8e-5& 0.0035$\pm$1.7e-5 &0.0011$\pm$1.0e-6\\\hline
            \hline
      Resnet-56 on CIFAR-10&\textbf{Epoch 1} & \textbf{Epoch 10} & \textbf{Epoch 25}& \textbf{Epoch 100}\\
      \hline
      \textbf{LR 0.3}&0.0380$\pm$5.9e-4 & 0.0131$\pm$4.5e-4& 0.0094$\pm$1.3e-5&0.0017$\pm$1.0e-5\\\hline
      \textbf{LR 0.15}&0.0084$\pm$5.2e-5& 0.0034$\pm$3.2e-5& 0.0020$\pm$7.2e-6&0.0013$\pm$6.7e-6\\\hline
      \hline
      VGG-11 on Tiny-ImageNet&\textbf{Epoch 1} & \textbf{Epoch 10} & \textbf{Epoch 25}& \textbf{Epoch 100}\\
      \hline
      \textbf{LR 0.5}&0.028$\pm$1.0e-3 & 0.213$\pm$1.5e-3& 0.187$\pm$1.9e-3&9.8e-5$\pm$2.0e-9\\\hline
      \textbf{LR 0.1}&0.0039$\pm$5.2e-5& 0.163$\pm$2.64e-5& 0.116$\pm$0.013&1.1e-5$\pm$3.6e-11\\\hline
    \end{tabular}
  \end{center}
\vspace{-20pt}
\end{table*}
\section{Learning Rate Schedule}
\label{section_lr_sch}
We observe from table \ref{tab:height_vgg} that the optimization oscillates at a lower height as training progresses (which is likely because SGD finds flatter regions as training progresses, see Figure \ref{fig:vgg11_cifar10_hessian_valacc}). As we discussed based on Figure \ref{fig:interpolation_vgg11_cifar10_epoch1}, the floor of the DNN valley is highly non-linear with many barriers. Based on these two observations, it seems that it should be advantageous for SGD to maintain a large height from the floor of the valley to facilitate further exploration without getting hindered by barriers as it may allow the optimization to find flatter regions. Hence, this line of thought suggests that we should increase the learning rate as training progresses (of course eventually it needs to be annealed for convergence to a minimum). \citet{smith2017cyclical, smith2017super} propose a cyclical learning rate (CLR) schedule which partially has this property. It involves linearly increasing the learning rate every iteration until a certain number of iterations, then similarly linearly reducing it, and repeat this process in a cycle. We now empirically show that multiple cycles of CLR are redundant, and simply increasing the learning rate until a certain point, and then annealing it leads to similar or better performance. Specifically, to rule out the need for cycles, as a null hypothesis, we increase the learning rate as in the first cycle of CLR, then keep it flat, then linearly anneal it (we call it the \textit{trapezoid schedule}). For fairness, we also plot the widely used step-wise learning rate annealing schedule. \textit{In our experiments, we find that methods which increase learning rate during training may be considered slightly better}. The learning curves are shown in figures \ref{fig:cyclic} in main text and \ref{fig:clr_resnet} in appendix (with other details). We leave an extensive study of learning rate schedule design based on the proposed guideline as future work.

We run the same experiment as described above on Resnet-56 with CIFAR-10 and it shows the rule for CLR, trapezoid schedule and SGD with stepwise annealing. Plots can be seen at figure \ref{fig:clr_resnet}. All schedules are tuned to their best performance with a hyperparameter grid search. For both Resnet-56 and VGG-11, we use batch size 100 for all models. The learning rate schedules are apparent from the figures themselves.

\section{Importance of SGD Noise Structure}
\label{appendix_hessian_covariance}
Here we derive in detail the relation between the Hessian and gradient covariance using the fact that for the negative log likelihood loss $\mathcal{L}_i(\theta) = -\log(p_i(\theta))$. Note we use the fact that for this particular loss function, $\frac{\partial \mathcal{L}_i(\theta)}{\partial p_i(\theta)} = -\frac{1}{p_i(\theta)}$, and $\frac{\partial^2 \mathcal{L}_i(\theta)}{\partial p_i(\theta)^2} = \frac{1}{p_i^2(\theta)}$, which yields $\frac{\partial^2 \mathcal{L}_i(\theta)}{\partial p_i(\theta)^2} = \left(\frac{\partial \mathcal{L}_i(\theta)}{\partial p_i(\theta)}\right)^2$.
\begin{align}
\mathbf{H}(\mathbf{\theta}) &= \frac{1}{N} \sum_{i=1}^{N} \frac{\partial^2 \mathcal{L}_i(\theta)}{\partial \theta^2} \\
&= \frac{1}{N} \sum_{i=1}^{N} \frac{\partial }{\partial \theta} \left( \frac{\partial \mathcal{L}_i(\theta)}{\partial p_i(\theta)} \cdot \frac{{\partial p_i(\theta)}}{\partial \theta} \right) \\
&= \frac{1}{N} \sum_{i=1}^{N} \frac{\partial^2 \mathcal{L}_i(\theta)}{\partial p_i(\theta)^2} \cdot \frac{\partial p_i(\theta)}{\partial \theta} \frac{\partial p_i(\theta)}{\partial \theta}^{T} \nonumber \\
&+ \frac{\partial \mathcal{L}_i(\theta)}{\partial p_i(\theta)} \cdot \frac{\partial^2 p_i(\theta)}{\partial \theta^2}\\
&= \frac{1}{N} \sum_{i=1}^{N} \left( \frac{\partial \mathcal{L}_i(\theta)}{\partial p_i(\theta)} \right)^2 \cdot \frac{\partial p_i(\theta)}{\partial \theta} \frac{\partial p_i(\theta)}{\partial \theta}^{T} \nonumber \\
&+ \frac{\partial \mathcal{L}_i(\theta)}{\partial p_i(\theta)} \cdot \frac{\partial^2 p_i(\theta)}{\partial \theta^2}\\
&= \frac{1}{N} \sum_{i=1}^{N} \frac{\partial \mathcal{L}_i(\theta)}{\partial \theta} \frac{\partial \mathcal{L}_i(\theta)}{\partial \theta}^{T} + \frac{\partial \mathcal{L}_i(\theta)}{\partial p_i(\theta)} \cdot \frac{\partial^2 p_i(\theta)}{\partial \theta^2}\\
&= \mathbf{C}(\theta)  + {\mathbf{\bar{g}}(\theta)}{\mathbf{\bar{g}}(\theta)}^T + \frac{1}{N} \sum_{i=1}^{N} \frac{\partial \mathcal{L}_i(\theta)}{\partial p_i(\theta)} \cdot \frac{\partial^2 p_i(\theta)}{\partial \theta^2}
\end{align}
where $\mathbf{\bar{g}}(\theta) = \frac{1}{N} \sum_{i=1}^{N} \frac{\partial \mathcal{L}_i(\theta)}{\partial \theta}$.
\section{Discussion}
\begin{figure}[ht]
\centering
\includegraphics[width=\columnwidth ,trim=0.in 0.3in 0.in 0.in,clip]{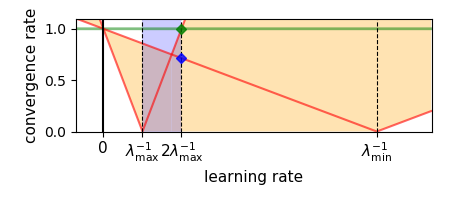}
\caption{Graph of $|\lambda_i \eta - 1|$: convergence rates of gradient descent on $\lambda_{\min}$-strongly
convex and $\lambda_{\max}$-smooth quadratic surfaces. The area which is shaded by orange contains possibly graphs of other eigenvalues of the hessian. In the range of learning rates shaded by blue, trajectory underdamps in the direction of the maximum eigenvalue. For a certain learning rate, while the trajectory oscillates in the direction of the maximum eigenvalue (green diamond), it overdamps in all the others (e.g. blue diamond - in a 'flat' direction).}
\label{fig:rates_of_gd_on_quadratics}
\vspace{-15pt}
\end{figure}

In the main text, we talk about converge in the quadratic setting depending on the value of learning rate relative to the largest eigenvalue of the Hessian. The convergence in this setting has been visualized in \ref{fig:rates_of_gd_on_quadratics}.
\twocolumn
\begin{figure}[!ht]
\begin{subfigure}[t]{1\columnwidth}
  \centering
  \includegraphics[scale=0.6,trim=0.in 0.35in 0.in 0.in,clip]{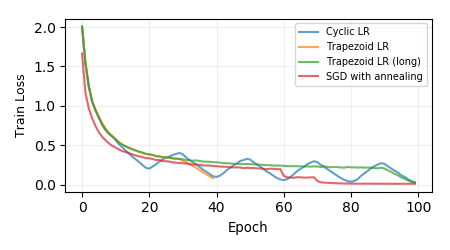}
\end{subfigure}\hspace{0.2\textwidth}
\begin{subfigure}[t]{1\columnwidth}
  \centering
  \includegraphics[scale=0.6,trim=0.in 0.35in 0.in 0.in,clip]{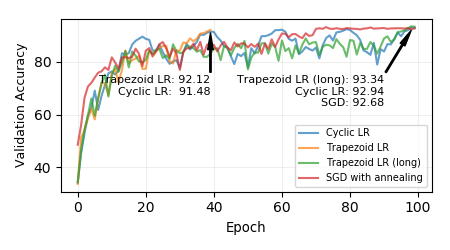}
\end{subfigure}\hspace{0.1\textwidth}
\begin{subfigure}[t]{1\columnwidth}
  \centering
  \includegraphics[scale=0.6,trim=0.in 0.35in 0.in 0.in,clip]{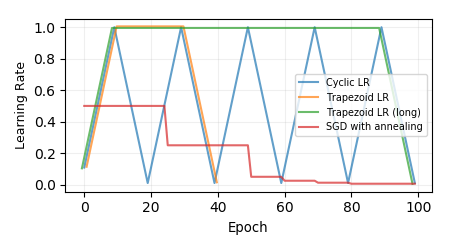}
\end{subfigure}\hspace{0.1\textwidth}
\caption{Plots for Resnet-56 on CIFAR-10 trained using Cyclic learning rate (CLR), SGD with stepsize annealing, and trapezoid schedule. Cycles in the CLR schedule are redundant, which is shown by the trapezoid schedule.}
\label{fig:clr_resnet}
\end{figure}

\begin{figure}[!ht]
\vspace{-5pt}
 \centering
  \includegraphics[scale=0.6,trim=0.in 0.37in 0.in 0.13in,clip]{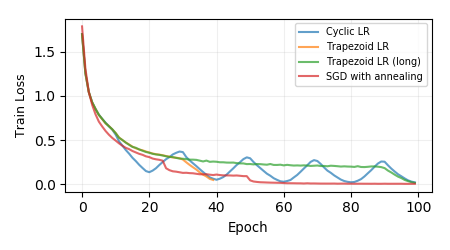}
  \includegraphics[scale=0.6,trim=0.in 0.37in 0.in 0.13in,clip]{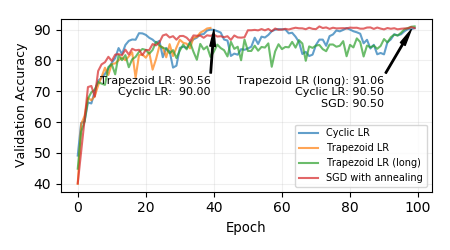}
  \includegraphics[scale=0.6,trim=0.in 0.17in 0.in 0.13in,clip]{figs/vgg_lr.png}
\captionof{figure}{Plots for VGG-11 on CIFAR-10 trained using Cyclic learning rate (CLR), SGD with stepwise annealing, and trapezoid schedule. Cycles in the CLR schedule are redundant, which is shown by the trapezoid schedule.}
\label{fig:cyclic}
\end{figure}

\begin{figure}
\centering
\includegraphics[width=\columnwidth ,trim=0.in 0.0in 0.in 0.in,clip]{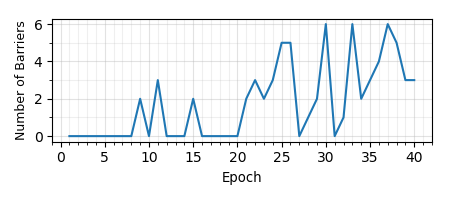}
\caption{Numbers of barriers found during training loss interpolation for every epoch(450 iterations) for VGG-11 on CIFAR-10. We say a barrier exists during a training update step if there exists a point between the parameters before and after an update which has a loss value higher than the loss at either points. Note that even for these barriers, their heights (defined by $\frac{\mathcal{L}(\mathbf{\theta}_t) + \mathbf{\theta}_{t+1}) - 2\mathcal{L}(\mathbf{\theta}_t^{\min})}{2})$ are substantially smaller compared with the value of loss at the corresponding iterations (not mentioned here), meaning they are not significant barriers.}
\label{fig:barrier}
\end{figure}



\begin{figure}[!ht]
 \centering
\begin{subfigure}[t]{1\columnwidth}
  \centering
  \includegraphics[scale=0.55,trim=0.in 0.35in 0.in 0.in,clip]{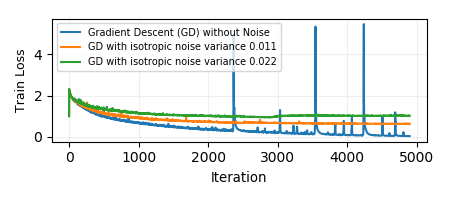}
\end{subfigure}\hspace{0.2\textwidth}
\begin{subfigure}[t]{1\columnwidth}
  \centering
  \includegraphics[scale=0.55,trim=0.in 0.35in 0.in 0.in,clip]{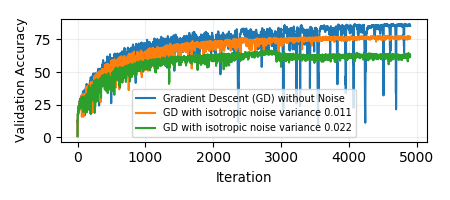}
\end{subfigure}\hspace{0.1\textwidth}
\begin{subfigure}[t]{1\columnwidth}
  \centering
  \includegraphics[scale=0.55,trim=0.in 0.35in 0.in 0.in,clip]{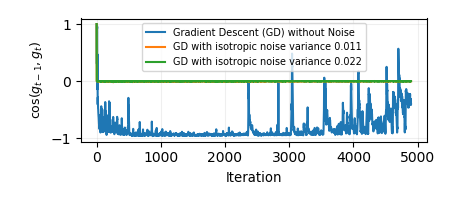}
\end{subfigure}\hspace{0.1\textwidth}
\begin{subfigure}[t]{1\columnwidth}
  \centering
  \includegraphics[scale=0.55,trim=0.in 0.35in 0.in 0.in,clip]{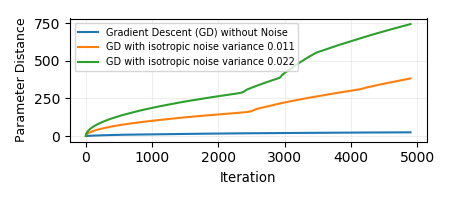}
\end{subfigure}\hspace{0.1\textwidth}
\caption{Plots for Resnet-56 trained by GD (without noise) and GD with artificial isotropic noise sampled from Gaussian distribution with different variances. Models trained using GD with added isotropic noise get stuck in terms of training loss and have worse validation performance compared with the model trained with GD.}
\label{fig:resnet_isotropic}
\end{figure}

\begin{figure}[!ht]
 \centering
\begin{subfigure}[t]{1\columnwidth}
  \centering
  \includegraphics[scale=0.55,trim=0.in 0.35in 0.in 0.in,clip]{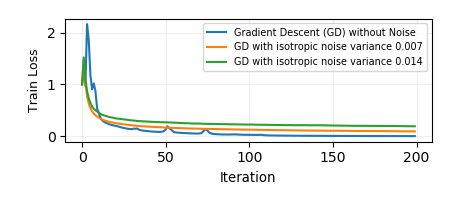}
\end{subfigure}\hspace{0.2\textwidth}
\begin{subfigure}[t]{1\columnwidth}
  \centering
  \includegraphics[scale=0.55,trim=0.in 0.35in 0.in 0.in,clip]{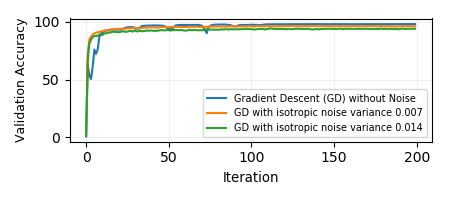}
\end{subfigure}\hspace{0.1\textwidth}
\begin{subfigure}[t]{1\columnwidth}
  \centering
  \includegraphics[scale=0.55,trim=0.in 0.35in 0.in 0.in,clip]{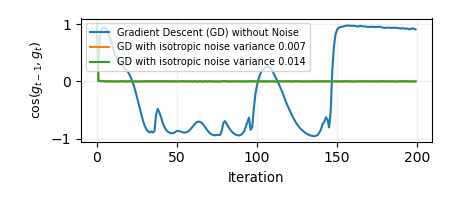}
\end{subfigure}\hspace{0.1\textwidth}
\begin{subfigure}[t]{1\columnwidth}
  \centering
  \includegraphics[scale=0.55,trim=0.in 0.35in 0.in 0.in,clip]{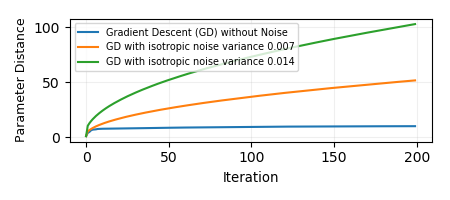}
\end{subfigure}\hspace{0.1\textwidth}
\caption{Plots for MLP trained by GD (without noise) and GD with artificial isotropic noise sampled from Gaussian distribution with different variances. Models trained using GD with added isotropic noise get stuck in terms of training loss and have worse validation performance compared with the model trained with GD.}
\label{fig:mlp_isotropic}
\end{figure}

